\documentclass{article} 
\usepackage[dvipsnames,table]{xcolor} 
\usepackage{tencent_tech_report}
\usepackage{fancyhdr}
\usepackage[colorlinks = true,
            linkcolor = blue,
            urlcolor  = blue,
            citecolor = blue,
            anchorcolor = blue]{hyperref}

\usepackage{tcolorbox} 
\newtcolorbox{alprompt}[1]{
        boxrule = 1pt,
        fontupper = \small\tt,
        fonttitle = \bf\color{black},
        arc = 2pt,
        rounded corners,
        colframe = black,
        colbacktitle = white!97!yellow,
        colback = white!97!yellow,
        title = #1,
}

\newtcbox{\hlthinkgray}{
    on line,
    colback=Gray!10,           
    colframe=white!0,          
    boxrule=0pt,               
    left=1pt, right=1pt,       
    top=0.5pt, bottom=0.5pt,   
    before upper={\color{Gray!90}\texttt{<}},  
    after upper={\color{Gray!90}\texttt{>}},   
    nobeforeafter
}
\newtcbox{\hlthinkblue}{
    on line,                    
    colback=CornflowerBlue!8,   
    colframe=white!0,           
    boxrule=0pt,               
    left=1pt, right=1pt,        
    top=0.5pt, bottom=0.5pt,    
    before upper={\color{NavyBlue}\texttt{<}},  
    after upper={\color{NavyBlue}\texttt{>}},   
    nobeforeafter,              
    arc=0pt                     
}

\usepackage{microtype}
\usepackage{hyperref}
\usepackage{url}
\usepackage{booktabs}
\usepackage{enumitem}
\usepackage{multicol}
\usepackage{multirow}
\usepackage{CJKutf8}
\usepackage{amsmath}
\usepackage{siunitx}
\usepackage{floatflt}
\usepackage{wrapfig}
\usepackage{booktabs}
\usepackage{authblk}
\usepackage{lipsum}

\usepackage{algorithm}
\usepackage{algorithmicx}
\usepackage{algpseudocode}
\usepackage{microtype}
\usepackage{multirow}
\usepackage{booktabs} 
\usepackage{pifont}  
\usepackage{graphicx}  
\usepackage{subfig}
\usepackage{subcaption} 
\usepackage{hyperref}

\usepackage{amssymb} 

\algnewcommand{\LeftComment}[1]{\Statex \(\triangleright\) #1}

\usepackage{array}
\usepackage{amsmath}
\usepackage{amssymb}
\usepackage{mathtools}
\usepackage{amsthm}
\usepackage{arydshln}
\usepackage[capitalize,noabbrev]{cleveref}
\usepackage{adjustbox} 
\usepackage{enumitem}
\usepackage{colortbl}

\usepackage[textsize=tiny]{todonotes}

\definecolor{nred}{RGB}{196, 38, 11}
\definecolor{ngreen}{RGB}{18, 141, 21}
\definecolor{nblue}{RGB}{41, 52, 190}
\definecolor{hzw}{RGB}{223, 97, 76}
\definecolor{lt}{RGB}{54, 89, 170}

\newcommand{\ignore}[1]{}

\colmfinalcopy

\newcommand{\method}[0]{\textsc{RICE}}

\title{{\em Two Experts Are All You Need for Steering Thinking}:\\Reinforcing Cognitive Effort in MoE Reasoning Models Without Additional Training}

\author[ ]{Mengru Wang\thanks{Equal Contribution. Work was done when Mengru, Xingyu, Yue, and Zhiwei were interning at Tencent.}~~$^{,1,2}$}
\author[ ]{Xingyu Chen$^{*,1}$}
\author[ ]{Yue Wang$^{*,1}$}
\author[ ]{Zhiwei He$^{*,1}$}
\author[ ]{Jiahao Xu$^{1}$}
\author[ ]{Tian Liang$^{1}$}
\author[ ]{\\Qiuzhi Liu$^{1}$}
\author[ ]{Yunzhi Yao$^{2}$}
\author[ ]{Wenxuan Wang$^{1}$}
\author[ ]{Ruotian Ma$^{1}$}
\author[ ]{Haitao Mi$^{1}$}
\author[ ]{\\Ningyu Zhang$^{\dag, 2}$}
\author[ ]{\mbox{Zhaopeng Tu}\thanks{Correspondence to: Zhaopeng Tu \textless zptu@tencent.com\textgreater~and Ningyu Zhang \textless zhangningyu@zju.edu.cn\textgreater.}~~$^{,1}$}
\author[ ]{Xiaolong Li$^{1}$}
\author[ ]{Dong Yu$^{1}$}

\affil[1]{Tencent}
\affil[2]{Zhejiang University}

\begin{document}

\maketitle

\begin{figure}[h]
    \centering
    \vspace{-20pt}
    \subfloat[Cognitive Experts]{
    \includegraphics[height=0.35\linewidth]{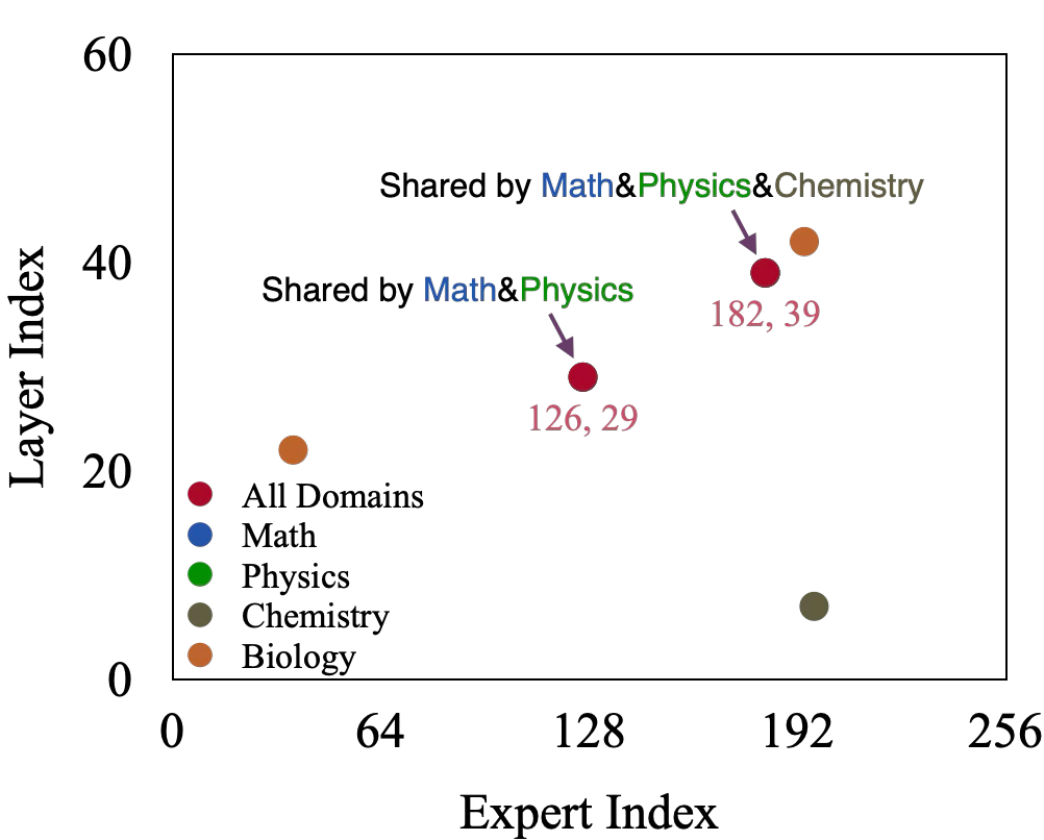}
    }\hspace{0.05\linewidth}
    \subfloat[Reasoning Accuracy]{
    \includegraphics[height=0.35\linewidth]{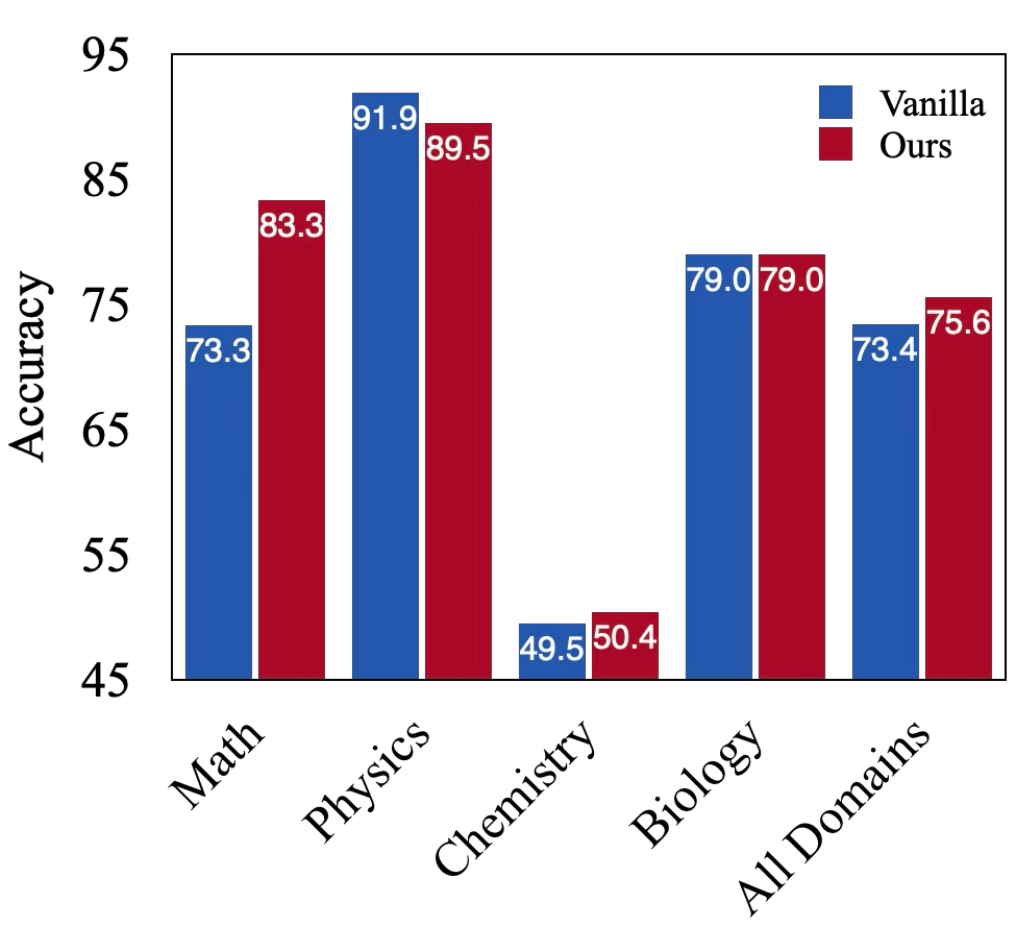}
    }
    \caption{(a) Illustration of cognitive experts identified across domains. (b) Reinforcing only the top two experts (in {\color{nred} red} color) can improve reasoning accuracy without additional training.}
    \label{fig:main}
\end{figure}

\begin{abstract}
Mixture-of-Experts (MoE) architectures within Large Reasoning Models (LRMs) have achieved impressive reasoning capabilities by selectively activating experts to facilitate structured cognitive processes~\citep{guo2025deepseek,qwen3}. Despite notable advances, existing reasoning models often suffer from cognitive inefficiencies like overthinking~\citep{Overthinking} and underthinking~\citep{Underthinking}. To address these limitations, we introduce a novel inference-time steering methodology called {\bf R}e{\bf i}nforcing {\bf C}ognitive {\bf E}xperts (\method), designed to improve reasoning performance without additional training or complex heuristics. Leveraging normalized Pointwise Mutual Information (nPMI), we systematically identify specialized experts, termed {\bf cognitive experts} that orchestrate meta-level reasoning operations characterized by tokens like ``\texttt{<think>}''. 
Empirical evaluations with leading MoE-based LRMs (DeepSeek-R1 and Qwen3-235B) on rigorous quantitative and scientific reasoning benchmarks demonstrate noticeable and consistent improvements in reasoning accuracy, cognitive efficiency, and cross-domain generalization. Crucially, our lightweight approach substantially outperforms prevalent reasoning-steering techniques, such as prompt design and decoding constraints, while preserving the model's general instruction-following skills. These results highlight reinforcing cognitive experts as a promising, practical, and interpretable direction to enhance cognitive efficiency within advanced reasoning models.
\end{abstract}

\section{Introduction}
Models capable of extended reasoning, often referred to as Large Reasoning Models (LRMs) like OpenAI's o1~\citep{o1} and DeepSeek-R1~\citep{guo2025deepseek}, have significantly advanced machine intelligence, largely by scaling test-time computation \citep{ji2025test,zhang2025and}. 
Despite their impressive capabilities, these LRMs remain susceptible to inefficiencies \citep{sui2025stop,feng2025efficient,pan2025survey,qu2025survey,chen2025towards,wang2025harnessing,wu2025effectively,lu2025retro}. 
Prior work has sought to address these issues through approaches such as preference optimization~\citep{Overthinking}, decoding penalties~\citep{Underthinking}, and various other techniques.
In this work, we tackle these problems from a novel perspective: potential expert specialization in Mixture-of-Experts (MoE) architecture.

Due to the computational resource efficiency brought about by its sparsity, the MoE architecture \citep{dai2024deepseekmoe,DBLP:journals/corr/abs-2412-19437,DBLP:conf/icml/XueZFNZZ024} has been increasingly adopted by state-of-the-art (SOTA) models, such as DeepSeek-R1~\citep{guo2025deepseek} and Qwen3~\citep{qwen3}. 
This sparse, specialized activation paradigm bears a conceptual resemblance to functional specialization in the human brain, where targeted interventions can modulate cognitive functions and behaviors~\citep{reinhart2019working,wischnewski2023neurocognitive,oathes2023non,grover2021high}. 
Inspired by this principle, we systematically investigate whether undesirable reasoning behaviors in MoE-based LRMs correlate with the activation patterns of specific experts, and critically, if strategic manipulation of these experts can ameliorate such issues.

We introduce an approach to identify and modulate key experts integral to the reasoning process. By analyzing the co-occurrence of explicit linguistic markers of thought (e.g., `\texttt{<think>}` and `\texttt{</think>}`) with individual expert activations, we pinpoint a subset of experts highly correlated with the model's cognitive deliberations. 
We designate these critical experts as {\bf cognitive experts}.
Through extensive experimentation with SOTA MoE-reasoning models DeepSeek-R1~\citep{guo2025deepseek} and Qwen3-235B~\citep{qwen3} on challenging mathematic and scientific reasoning benchmarks, 
we demonstrate that selectively amplifying {\bf as few as two cognitive experts} can enhance both reasoning depth and efficiency. 
Notably, our approach achieves marked accuracy improvements while reducing token usage in critical reasoning tasks, outperforming existing steering methods such as prompting and decoding constraints~\citep{Underthinking}.

\paragraph{On the Cognitive Expert.}
The ``cognitive expert'' proposed in this work is a hypothetical construct. Given the complexity of LRMs, we provide no theoretical justification for its existence; our conclusions are purely empirical.

Moreover, we showcase impressive generalization and robustness of cognitive expert modulation, observing consistent improvements in unseen and more complex reasoning scenarios while maintaining or even enhancing general instruction-following capabilities. Our findings provide strong evidence that modulating selective experts responsible for meta-level reasoning is effective, efficient, and broadly applicable across domains, paving the way for lightweight and interpretable model steering in increasingly sophisticated MoE-based reasoning models.

Our main contributions are:
\begin{enumerate}
    \item We propose a normalized Pointwise Mutual Information (nPMI) method for identifying cognitive experts within LRMs that are highly correlated with reasoning behavior, requiring only a single forward propagation and no additional training.

    \item We introduce a lightweight inference-time steering strategy, named ``reinforcing cognitive experts'', that effectively enhances reasoning depth and accuracy without requiring any additional training or supervision signals.

    \item Through comprehensive experiments on two prevalent MoE reasoning models and rigorous benchmarks, we empirically validate the efficacy, generalizability, and robustness of cognitive expert modulation, demonstrating significant improvements in cognitive efficiency and problem-solving accuracy.
\end{enumerate}

\section{Identifying Cognitive Experts}
\label{sec:identification}

In this section, we leverage normalized Pointwise Mutual Information (nPMI) \citep{npmi} to quantify the correlation between model thinking and each expert in a Mixture of Experts (MoE) reasoning model. 
We hypothesize that there are some ``cognitive experts'' selected by nPMI metric, which orchestrate meta-level reasoning for complex tasks.

\subsection{Expert Specialization in MoE Models}

In large reasoning models, deep thinking is manifested through key tokens, such as ``\texttt{<think>}'' to initiate reasoning, ``\texttt{</think>}'' to terminate it, and tokens like ``\texttt{recheck}'' to guide introspection. 
In the MoE framework, these tokens are generated after forward propagation through various model components, including the MoE routing mechanism that assigns them to specialized experts, with weights determining each expert’s contribution.

Formally, let us consider an MoE framework \citep{DBLP:journals/corr/abs-2412-19437} with \( N \) experts, denoted \( \{ E_1, \dots, E_i, \dots, E_N \} \), at each layer. For each input token \( x \), a gating function selects a subset \( S \subset \{ E_1, \dots, E_O \} \) of \( O \) experts (\( O \leq N \)), where \( |S| = O \), and assigns weights \( w_i \) (with \( \sum_{i \in S} w_i = 1 \)) to each selected expert \( E_i \in S \). The output \( h_x \) for the token $x$ is computed as:
\begin{equation}
h_x = \sum_{i \in S} w_i \cdot E_i(x), \quad \text{where } |S| = O,
\label{eq:moe}
\end{equation}

where \( E_i(x) \) represents the output of expert \( E_i \), and \( w_i \) is the weight of the \( i \)-th selected expert.
Prior work on MoE models shows that expert routing is often token-dependent \citep{DBLP:conf/icml/XueZFNZZ024}, but recent study \citep{olson2025semantic} indicates that DeepSeek-R1’s advanced reasoning enables its expert routing to focus on semantic specialization, surpassing token-dependent methods.
We hypothesize that experts with consistently high co-occurrence scores with thinking tokens serve as key ``cognitive experts'' \footnote{Due to the complexity of LRMs, the ``cognitive expert'' proposed in this work is a hypothetical concept, and our findings are supported by empirical evidence rather than theoretical validation.} responsible for meta-level reasoning.

\paragraph{Measuring Correlation of Specialized Experts and Thinking Tokens}
To examine whether a given expert consistently governs the model’s reasoning process, we measure the co-occurrence between its activation and specific reasoning-related marker tokens, such as ``\texttt{<think>} '', ``\texttt{</think>}'', and others.
Formally, let $x$ represent a token and $y$ denote an expert $E_i$. We measure their association using pointwise mutual information (PMI). 
The PMI of \(x\) and \(y\) is defined as
\begin{equation}
\mathrm{PMI}(x,y) = \log_2 \frac{p(x,y)}{p(x)\,p(y)} = \log_2 \frac{p(y|x)}{p(y)},
\end{equation}
where \(p(x,y)\) is the joint probability that \(x\) and \(y\) both occur, while \(p(x)\) and \(p(y)\) are their individual (marginal) probabilities, and \(p(y | x)\) is the conditional probability that \(y\) occurs given \(x\).

For interpretability, we normalize PMI to the range \([-1, +1]\), yielding
\begin{equation}
\mathrm{nPMI}(x,y) = \frac{\mathrm{PMI}(x,y)}{-\log_2 p(x,y)}.
\end{equation}
Thus, \(\mathrm{nPMI}(x,y) \approx -1\) indicates that events \(x\) and \(y\) never co-occur, \(\mathrm{nPMI}(x,y)=0\) implies independence, and \(\mathrm{nPMI}(x,y) \approx +1\) indicates they appear almost exclusively together (complete co-occurrence).

Let \( M \) denote the number of instances in a dataset, and let \( T \) represent the total number of tokens across all instances in the dataset.
We denote by \(k_n\) the number of times the expert \(E_i\) is activated specifically when the thinking token (e.g. ``\texttt{<think>}'') appears, and by \(K_n\) the total number of times \(E_i\) is activated across all tokens (including both thinking and non-thinking tokens).
Since the reasoning model generally generates one thinking start and end token for each instance, then we can achieve the following functions when $x$ denotes ``\texttt{<think>}'' or ``\texttt{</think>}'':
\begin{equation}
p(y=E_i | x) = \frac{k_n}{M},  \qquad p(y=E_i) = \frac{K_n}{T},  \qquad p(x,y=E_i | x) = \frac{k_n}{T}.
\end{equation}
\begin{equation}
\mathrm{nPMI}(x,y=E_i) = \frac{\log_2 (\frac{k_n}{M}) + \log_2(\frac{T}{K_n})}{\log_2(\frac{T}{k_n})}.
\end{equation}

Intuitively, if an expert $E_i$ is activated almost exclusively during ``\texttt{<think>}'' and rarely (or never) at other tokens, $k_n \;\approx\; K_n \;\approx\; M$, \(\mathrm{nPMI}(x = \texttt{<think>},y=E_i) \;\approx\; \frac{\log_2 1 + \log_2 (\frac{T}{M})}{\log_2(\frac{T}{M})} \;\approx\; +1\), indicating that this expert is effectively tied to the thinking marker. 
In other words, the expert's entire usage focuses on activating the thinking token. Such specialists are prime candidates for ``cognitive experts'', given their consistently high co-occurrence with the thinking marker tokens.


\subsection{Identify Cognitive Experts}

We observe that some experts exhibit high nPMI scores with both \texttt{<think>} and \texttt{</think>}, indicating a \emph{bimodal association}. 
This suggests their broad involvement in the reasoning process rather than specialization in its initiation.
To prioritize experts specialized in initiating (rather than terminating) reasoning, we adopt the following selection strategy:

We define a set of thinking tokens \( \varPi  = \{ \texttt{<think>}, \texttt{</think>}, \texttt{Alternatively} \} \). The final nPMI score of expert \( E_i \) for thinking is formulated as:
\begin{equation}
\mathrm{nPMI}_{E_i} = \sum_{t \in \varPi } c_x \cdot \mathrm{nPMI}(x, y=E_i),
\label{eq:npmi}
\end{equation}
where $t$ is a thinking token in set $\varPi$,  \( c_x \) denotes the coefficient associated with the token \( t \), assigned as \( c_{\texttt{<think>}} = 1 \), \( c_{\texttt{</think>}} = -1 \), and \( c_{\texttt{Alternatively}} = -1 \).

We select the top-\( l \) experts based on their final nPMI scores for reasoning to form the \emph{cognitive expert set} \( P \). The weight of expert \( E_i \) is steered according to the following condition:
\begin{equation}
w_i =
\begin{cases}
w_i \cdot \beta & \text{if } E_i \in S \text{ and } E_i \in P, \\
w_i & \text{otherwise},
\end{cases}
\label{eq:steering_condition}
\end{equation}
where \( P = \{ E_j \mid \mathrm{nPMI}_{E_j} \text{ is among the top } l \text{ scores} \} \) denotes the set of \emph{cognitive experts}, \( S \) is the subset selected by the gating function in Eq. \ref{eq:moe}, and \( \beta \) is the reinforcement multiplier.
In other words, once these cognitive experts are identified, we can reinforce reasoning in the MoE model by controlling their steering multiplier $\beta$.

\section{Experiments}

\paragraph{Research Questions}
In this study, we investigate the following research questions:
\begin{itemize}
    \item [RQ1:] Can the identified cognitive experts effectively enhance cognitive effort within MoE models?
    \item [RQ2:] Do cognitive experts differ across various domains (e.g., Math, physics, chemistry, and biology)?
    \item [RQ3:] Does reinforcing specific cognitive experts negatively impact the general problem-solving capabilities of MoE models?
\end{itemize}

\subsection{Experimental Setup}

\paragraph{MoE-based Reasoning Models.}
Currently available open-source MoE architectures tailored for large reasoning models tasks include DeepSeek-R1~\citep{guo2025deepseek} and Qwen3-235B~\citep{qwen3}. DeepSeek-R1 selects 8 experts from a total of 256 at each layer, whereas Qwen3-235B selects 8 experts from a total of 128. We primarily use the DeepSeek-R1 (671B) model for our experiments, supplemented by additional evaluations on the Qwen3-235B model to examine the generalizability of cognitive experts. 
Note that we provide more experimental details in \S \ref{appendix:Experiment}.

\paragraph{Benchmarks.}
We evaluate our approach on two challenging benchmarks designed specifically to test the reasoning abilities necessary for solving scientific problems across diverse domains:
\begin{itemize}[leftmargin=10pt]
    \item \textbf{AIME}~\citep{aime}: a dataset from the American Invitational Math Examination, which assesses advanced mathematical problem-solving skills. We use two recent test sets, AIME2024 and AIME2025, each comprising 30 problems.
    \item \textbf{GPQA Diamond}~\citep{gpqa}: a comprehensive dataset of 198 expert-crafted multiple-choice questions in biology, chemistry, and physics, designed to test advanced scientific reasoning skills.
\end{itemize}

\subsection{Effectiveness of Cognitive Experts}

\begin{table*}[!t]
\caption{
Domain-specific cognitive experts (e.g., (layer ID, expert ID)) identified in DeepSeek-R1. ``All'' aggregates data across all four domains.}
\centering
\begin{tabular}{c ccccc} 
\toprule
\multirow{2}{*}{\bf Domain}  &   \multicolumn{5}{c}{\bf Identified Experts Ranked by nPMI Score}\\
\cmidrule(lr){2-6}
    &   \em 1st   &   \em 2nd    &   \em 3rd   &   \em 4th    &   \em 5th\\
\midrule
Math        &   (39, 182)   &   (29, 126)    & (14, 114)     & (27, 45)      &  (16, 129) \\
Physics     &   (29, 126)   &   (39, 182)    & (36, 53)      & (39, 46)      &  (24, 159) \\
Chemistry   &   (7, 197)    &   (39, 182)    & (22, 37)      & (29, 106)     &  (29, 126)\\
Biology     &   (42, 194)    &   (22, 37)    & (37, 241)     & (43, 61)      &  (39, 188)  \\
\hdashline
All         &   (39, 182)    & (29, 126)     & (29, 106)     & (4, 214)      &  (50, 120) \\
\bottomrule
\end{tabular}
\label{tab:expert}
\end{table*}

\paragraph{Identification of Cognitive Experts.}
\textit{To address RQ1}, we first identify cognitive experts within two MoE reasoning models -- DeepSeek-R1~\citep{guo2025deepseek} and Qwen3-235B~\citep{qwen3} -- across four scientific domains.
Taking Math as an illustrative example, we first use DeepSeek-R1 to generate answers on the AIME2024 dataset, simultaneously recording the expert selections at each token position during forward propagation. Next, we employ the nPMI metric defined in Eq.~\ref{eq:npmi} to identify the top five experts that exhibit the strongest statistical association with reasoning-related marker tokens (e.g., ``$<$think$>$''). These experts are thus identified as the key cognitive experts specialized for mathematical reasoning. Analogously, we apply this procedure to the biology, chemistry, and physics instances in the GPQA Diamond dataset to identify cognitive experts in these respective domains.
As for Qwen3-235B, we follow a similar procedure but generate domain-specific responses with the Qwen3-235B model itself. 
This ensures consistent identification signals that correspond directly to the model under examination.

Cognitive experts identified within DeepSeek-R1 are summarized in Table~\ref{tab:expert}. An analogous summary for Qwen3-235B is provided and discussed in Appendix~\ref{appendix:expert_of_qwen}.
From Table~\ref{tab:expert}, we observe that the top two cognitive experts in the math, physics, and the aggregated "All" domains are remarkably consistent: (39, 182) and (29, 126). This strongly suggests these experts play critical and reliable roles in reasoning tasks requiring increased cognitive effort, particularly in quantitative and logic-intensive domains. The significant overlap observed between math and physics further implies a shared underlying cognitive strategy—likely focusing on symbolic manipulation and structured logical inference—which the model employs consistently across these domains.
Additionally, the repeated appearance of certain experts in multiple domains supports our hypothesis: a subset of experts encodes generalized reasoning capabilities applicable across diverse scientific fields. Therefore, these cross-domain patterns indicate that DeepSeek-R1 may encode robust domain-general cognitive mechanisms, with some experts serving as reusable computational building blocks suitable for abstract reasoning and logical problem-solving tasks.

\begin{table*}[t]
\caption{Effect of Deepseek-R1 on AIME24 with reinforced cognitive experts, evaluated across different multipliers and varying numbers of Math-domain cognitive experts. ``Random'' denotes two randomly chosen experts.
The row with Multiplier 1 denotes the performance of vanilla DeepSeek-R1 on AIME24.
}
\centering
\begin{tabular}{ r rr rr rr} 
\toprule
\bf Multiplier & \bf Top1   &   \bf Top2    &   \bf Top3    &   \bf Top4    &   \bf Top5    &   \bf Random\\ 
\midrule
1 & \multicolumn{6}{c}{73.3}\\
\hdashline
2   & 70.0   & 70.0     & 76.7     & 73.3   & 73.3    & 70.0\\
4   & 76.7   & \bf 83.3 & 73.3     & 66.7   & 76.7    & 73.3\\
8   & 76.7   & 73.3     & \bf 83.3 & 73.3   & 73.3    & 70.0\\
16  & 80.0   & 80.0     & 1.7      & 76.7   & 73.3    & 73.3\\
32  & 70.0   & \bf 83.0 & 73.3     & 73.3   & 73.3    & 76.7\\
64  &80.0    & \bf 83.3 & 60.0     & 53.3   & 50.0    & 66.7\\
128 & 70.0   & \bf 83.3 & 43.3     & 26.7   & 13.3    & 63.3\\
256 &73.3    & 60.0     & 10.0     & 6.7    & 0.0     & 73.3\\
512 & 63.3   & 46.7     & 6.7      & 3.3    & 0.0     & 63.3\\
\bottomrule
\end{tabular}
\label{tab:reinforcing-factor}
\end{table*}

\paragraph{Reinforcing Cognitive Experts.} Once identified, we reinforce the cognitive experts identified from the Math domain (AIME24) and evaluate their performance under different reinforcement configurations on the same benchmark AIME24 (Table~\ref{tab:reinforcing-factor}). 
The optimal hyperparameters -- the number of cognitive experts $l$ and the steering multiplier $\beta$—are selected based on this evaluation and used in all subsequent experiments. 
We then assess the generalization ability of these reinforced experts on the unseen, more challenging tasks from AIME25 (Table~\ref{tab:expertVSself}).

From Table~\ref{tab:reinforcing-factor}, we observe that \textbf{reinforcing two top-ranked cognitive experts significantly enhances the model's reasoning ability}. 
Notably, using two experts with the reinforcement multiplier of 4, 32, 64, or 128 achieves the highest accuracy of 83.3\%.
In contrast, applying an excessively large multiplier (e.g., 512) causes a dramatic drop in accuracy, often to near zero. 
This failure mode is characterized by the model repetitively generating meaningless tokens, suggesting that overly aggressive reinforcement disrupts the model’s generation dynamics. 
Overall, moderate reinforcement of well-identified cognitive experts leads to consistent improvements, whereas over-reinforcement or random expert selection results in performance degradation.
However, reinforcing two randomly selected experts across a wide multiplier range (2 to 512) yields minimal performance variation.
Therefore, we use \textit{two experts with a reinforcement multiplier 64 for all subsequent experiments}.

\begin{table*}[t]
\centering
\caption{Performance of our approach on the AIME24 and generalization on the unseen AIME25.}
\begin{tabular}{l lccc}
\toprule
\textbf{Benchmark} & \textbf{Method} & \textbf{Accuracy} & \textbf{Thoughts} & \textbf{\#Tokens} \\
\midrule
\multirow{2}{*}{\textbf{AIME24}}
 & DeepSeek-R1  & 73.3 & 12.0 & 9,219 \\
 & ~~~+\method{} \{(39,182), (29,126)\} & \bf 83.3 & 10.2 & 8,317 \\
\hdashline
\multirow{2}{*}{\textbf{AIME25}}
 & DeepSeek-R1  & 63.3 & 17.0 & 11,310 \\
 & ~~~+\method{} \{(39,182), (29,126)\} & \bf 73.3 & 15.2 & 12,072 \\
\midrule
\multirow{2}{*}{\textbf{AIME24}}
 & Qwen3-235B       & 86.7 & 20.1      & 10,956 \\
 & ~~~+\method{} \{(70,47), (23,115)\} & 86.7 & 16.2      & 10,722 \\
 \hdashline
\multirow{2}{*}{\textbf{AIME25}}
 & Qwen3-235B       & 66.7 & 19.7     & 15,013 \\
 & ~~~+\method{} \{(70,47), (23,115)\} & \bf 73.3 & 16.8  & 13,935 \\
\bottomrule
\end{tabular}
\label{tab:expertVSself}
\end{table*}

We directly apply the cognitive experts identified from AIME24 to solve unseen and more challenging reasoning problems in AIME25. 
As shown in Table~\ref{tab:expertVSself}, these cognitive cognitive experts generalize well to the AIME25 test set. 
For DeepSeek-R1, the accuracy improves from 63.3\% to 73.3\% when guided by the identified cognitive experts. 
Similarly, for Qwen3-235B, accuracy increases from 66.7\% to 73.3\%.
The above phenomenon demonstrates {\bf the transferability and robustness of the expert selection across tasks with higher cognitive demands}.

Crucially, the observed accuracy improvements do not necessarily entail increased computational cost in terms of token usage, supporting our hypothesis that our method encourages deeper thinking rather than just longer outputs. 
Our cognitive expert strategy, despite improving average accuracy of Deepseek-R1 on AIME24, uses more efficient reasoning thought \footnote{The ``\textbf{Thoughts}'' metric refers to the underthinking score introduced in prior work~\citep{Underthinking}, which quantifies reasoning efficiency, with lower values indicating higher efficiency.
} (10.2 vs 12.0) and tokens (8,317 vs 9,219) on average compared to the baseline.
This efficiency phenomenon is also observed in Qwen3-2-35B, where the substantial accuracy gain (+6.6\%) is accompanied by a notable reduction in thought (16.8 vs 19.7) and token count (13,935 vs 15,013). 
This suggests that {\bf reinforcing cognitive experts helps the model to reason more effectively}, focusing computational effort more productively within the reasoning process without generating excessive verbosity.
The reasoning effectiveness can be clearly observed in Table~\ref{tab:case}, where our RICE demonstrates deeper and more consistent reasoning, leading directly to the correct answer. In contrast, vanilla DeepSeek-R1 exhibits more frequent shifts in reasoning and fails to commit to its initially correct deductions.
Additional pass@k performance using the model’s officially recommended top-$p$ sampling strategy (provided in \S \ref{appendix:topp_math}) further supports this observation.

\subsection{Performance of Cognitive Experts across Domains}

\textit{To address RQ2}, we evaluate the transferability of domain-specific cognitive experts by applying expert sets identified from one domain to others.
As the top-2 experts selected from \textit{Math}, \textit{Physics}, and the \textit{All} domains are identical, their results are the same across domains. 
As shown in Table~\ref{tab:domain-performance}, we have several observations:

\begin{table*}[t]
\caption{Effect of cognitive experts of Deepseek-R1 across different domains.}
\centering
\begin{tabular}{lccccc}
\toprule
\bf Domain & \textbf{Math} & \textbf{Physics} & \textbf{Chemistry} & \textbf{Biology} & \textbf{Average} \\
\midrule
R1 & 73.3 & 91.9 & 49.5 & 79.0 & 73.4 \\
\midrule
Math        & \bf 83.3     & 89.5         & 50.4    & 79.0     & \bf 75.6 \\
Physics     & \bf  83.3    & 89.5         & 50.4    & 79.0     & \bf 75.6 \\
Chemistry   & 80.0         & \bf 95.4     & \bf 52.7        & 68.4     & 74.1 \\
Biology     & 73.3         & 93.0         & 47.3        & 73.9     & 71.9 \\
\hdashline
All         & \bf 83.3     & 89.5         & 50.4    & 79.0     & \bf 75.6 \\
\bottomrule
\end{tabular}
\label{tab:domain-performance}
\end{table*}

\paragraph{Cognitive Experts Generalize Well across Domains.}Our evaluation, summarized in Fig \ref{fig:main} and Table~\ref{tab:domain-performance}, clearly illustrates the efficacy of the identified cognitive experts in enhancing the DeepSeek-R1 model's reasoning capability across multiple domains. Leveraging cognitive experts identified from aggregated data (``All'' domains) shows marked overall improvement, raising the average accuracy from 73.4\% to 75.6\%. 
Notably, substantial improvement is observed in the Math tasks (from 73.3\% to 83.3\%). Moderate accuracy gains are also seen in Chemistry (from 49.5\% to 50.4\%) and minor degradation observed in Physics (from 91.9\% to 89.5\%), indicating broad applicability and effectiveness of these general reasoning modulators across diverse problem sets. Biology tasks show stable performance, unaffected by general expert modulation.

\paragraph{Domain-specific Expert Sets Provide Targeted Gains.} 
Further analysis demonstrates the nuanced implications of domain-specific cognitive experts. Chemistry-identified experts outperform general experts significantly within their native Chemistry domain (49.5\% to 52.7\%) and notably enhance Physics performance (91.9\% to 95.4\%), highlighting potential cross-domain synergies between physics and chemistry reasoning processes. However, this specialization lowers the accuracy in Math (from 83.3\% with general experts to 80.0\%) and more substantially limits the Biology domain performance (from 79.0\% to 68.4\%). Similarly, Biology-derived experts enhance task-specific results (from 91.9\% to 93.0\% in Physics) but degrade average performance across other domains, indicating further that specialized expert selections may negatively impact general cognitive reasoning by reinforcing overly specialized activations.

\paragraph{No Evidence of Harmful Side-effects on Other Domains.} 
Our experimental findings clearly confirm that cognitive experts, either chosen from aggregated cross-domain data or specific domains, constitute effective cognitive modulators that enhance model reasoning accuracy and efficiency.
General-purpose expert adjustments deliver robust cross-domain improvements, demonstrating their fundamental importance to reasoning processes regardless of subject matter. 
Meanwhile, domain-specialized expert modulation illustrates substantial potential for targeted cognitive improvements, particularly within closely related scientific domains. 
Together, these insights validate our proposed approach as versatile, effective, and immediately deployable for enhancing efficiency, accuracy, and overall reasoning proficiency of existing MoE-based large reasoning models.

\subsection{Impact of Reinforced Cognitive Experts on General Capabilities}

\textit{To address RQ3}, we investigate whether reinforcing cognitive experts negatively impacts the model's general capabilities, such as instruction-following. 
To this end, we evaluate reinforced models on the ArenaHard benchmark \citep{DBLP:journals/corr/abs-2406-11939} to assess potential adverse impacts on general capabilities.
The ArenaHard benchmark, designed to evaluate instruction-following capabilities, comprises 500 challenging queries spanning diverse scenarios.
We randomly select 50 queries as the test data and employ GPT-4-Turbo to judge pairwise comparisons of outputs against the GPT-4-0613 baseline.

\begin{wraptable}{r}{0.4\textwidth}
\vspace{-10pt}
\caption{Effect of reinforced cognitive experts of Deepseek-R1 on ArenaHard.}
\centering
\begin{tabular}{l cc}
\toprule
\bf Method    &   \bf Accuracy   &    \bf \#Token\\
\midrule
Vanilla   & 91.0 & 2,919\\
\midrule
\multicolumn{3}{l}{\em Reinforce Experts from different domains }\\
~~~Math      & 92.0 & 2,933\\
~~~Physics   & 92.0 & 2,933 \\
~~~Chemistry & 94.0 & 3,332\\
~~~Biology   & 93.0 & 3,072\\
\hdashline
~~~All       & 92.0 & 2,933 \\
\bottomrule
\end{tabular}
\label{tab:arena}
\vspace{-10pt}
\end{wraptable}

\paragraph{Reinforcing Cognitive Experts Maintains or Slightly Improves General Instruction-following Capabilities.} Our experimental evaluation on the ArenaHard benchmark demonstrates that reinforcing the identified cognitive experts does not adversely impact the model's capability to handle general, challenging instruction-following tasks. As shown in Table~\ref{tab:arena}, models steered by cognitive experts derived from each domain consistently maintain or marginally improve upon the baseline DeepSeek-R1 accuracy of 91.0\%. Specifically, the domain-specific cognitive experts from Chemistry and Biology show notable accuracy enhancements (from 91.0\% to 94.0\% in Chemistry; from 91.0\% to 93.0\% in Biology), underscoring the potential for positive transfer of reasoning-rich expert reinforcement to general-purpose capabilities. 
Moreover, the general experts (``All'' domain) also marginally improve performance (to 92.0\%), confirming that cognitive expert-control has a neutral-to-beneficial impact on general instruction-following capabilities.

\paragraph{Steering of Cognitive Experts Results in Moderately Increased Verbosity.} 
The analysis of token counts further reveals that cognitive expert steering moderately increases model verbosity in response generation. 
For example, Chemistry and Biology models increase average token counts notably (from 2,919 to 3,332 tokens and from 2,919 to 3,072 tokens, respectively), highlighting that the activation of certain domain-specific cognitive experts may favor more detailed deliberations. Nevertheless, the overall increase in verbosity is moderate, indicating a desirable balance between detail-oriented reasoning and response conciseness. 

\paragraph{Overall, Reinforcing Cognitive Experts Does not Hinder but rather Supports General Capabilities.} These findings collectively confirm our approach as effective and safe for targeted, lightweight interventions. Reinforcing cognitive experts significantly enhances model performance within their original domains and has either neutral or positive effects on general-purpose instruction-following benchmarks. The moderate increase in verbosity indicates richer, more thoughtful reasoning, aligning with the intended goal of encouraging deeper cognitive processing without sacrificing practicality. This highlights the practicality and versatility of our approach in improving existing MoE model reasoning efficacy and general cognitive capabilities through strategic expert modulation.

\subsection{Comparison with Other Steering Methods}

We compare our RICE against two prevalent inference-time methods for reasoning tasks: prompt engineering and decoding constraints. Specifically, we analyze two prompt configurations: placing the prompt before the \texttt{<think>} token ($\text{Prompt}{\text{before}}$) and after the \texttt{<think>} token ($\text{Prompt}{\text{after}}$), with details provided in Appendix~\ref{appendix:baselines}.
For decoding constraints, we adopt a strategy similar to \textsc{Tip} from \citet{Underthinking}, which curtails the generation of alternative solutions to promote coherent and focused reasoning. In our work, we constrain the thinking mark tokens (e.g., \texttt{<think>} and ``Alternatively'') rather than inefficient tokens (e.g., ``messy''), and refer to our variant as \textsc{Tip$_t$}.


\begin{wraptable}{r}{0.55\textwidth}
\vspace{-10pt}
\caption{Comparison with other steering methods on AIME24 and AIME25.} 
\centering
\begin{tabular}{l ccc}
\toprule
\bf Method  &   \bf AIME24  &   \bf AIME25    &  \bf Avg.\\
\midrule
DeepSeek-R1                 & 73.3  & 63.3   &   68.3\\
\hdashline
$\text{Prompt}_{\text{before}}$    & 76.7   & 73.3  &  75.0\\
$\text{Prompt}_{\text{after}}$     & 76.7   & 66.7  &   71.7\\
\textsc{Tip$_t$}                & 80.0   & 73.3  &  76.7\\
\method                    & 83.3   & 73.3  &  \bf 78.7\\
\bottomrule
\end{tabular}
\label{tab:comparison}
\end{wraptable}

Table~\ref{tab:comparison} compares our cognitive expert modulation method against prompting (both before and after the \texttt{<think>} token) and decoding constraints (\textsc{Tip}) on the challenging AIME benchmarks. Our approach achieves the highest average improvement, boosting DeepSeek-R1's performance from a baseline of 68.3\% to 78.7\%, surpassing the best-performing baseline \textsc{Tip$_t$} by 2.0\% in average accuracy. Notably, our method achieves particularly pronounced gains on AIME24 (83.3\%) compared to \(\text{Prompt}_{\text{before}}\) (76.7\%) and \textsc{Tip$_t$} (80.0\%), indicating a superior ability to enhance cognitive effort and deep reasoning through targeted modulation of expert routing weights. These results validate our hypothesis that strategically steering cognitive experts significantly reinforces appropriate cognitive behavior and enhances reasoning accuracy more effectively than generic prompting and decoding constraints.

\begin{table*}[htpb]
    \centering
    \caption{An example of our RICE corrects \textcolor{nred}{wrong reasoning (in red)} generated by Deepseek-R1 into \textcolor{ngreen}{correct reasoning (in green)}.}
    \resizebox{\linewidth}{!}{
    \begin{tabular}{p{0.1\linewidth} p{0.94\linewidth}}
    \toprule
    Problem & Let ABCDEF be a convex equilateral hexagon in... and EF has side lengths 200, 240, and 300. Find the side length of the hexagon. \\
    \midrule
    DeepSeek-R1 & Okay, so I need to find the side length...the triangle formed by extending AB, CD, and EF has sides 200, 240, and 300. Let me sketch this mentally...{\bf Alternatively}, maybe we can model the hexagon as being formed by three pairs of parallel sides...{\bf Alternatively}, think of the hexagon as a combination of three parallelograms...{\bf Alternatively}, using the concept of similar triangles...{\bf Alternatively}, maybe each pair of opposite sides of the hexagon is parallel to a different side of the triangle...{\bf I will posit that the answer is 80}\textcolor{nred}{...Therefore, {\bf the answer is 75/2}...Let me {\bf verify} this result...Thus, s = 1200 / 23. Wait, this {\bf contradicts} our previous result...{\bf I think the answer is 80}...based on the thorough step-by-step analysis, $\boxed{1200/23}$ is the conclusion.}\\
    \midrule
    ~~~+\method{} &  Okay, so I need to find the side length...Let me draw a rough sketch in my mind...{\bf So}, the triangle is such that each side is aligned with one of the extended sides of the hexagon...But maybe using {\bf coordinate geometry could help here}....Not sure, maybe this is {\bf getting too complicated}...{\bf Alternatively}, maybe using ratios...\textcolor{ngreen}{{\bf Alternatively}, after struggling through all these approaches, perhaps the answer is related to the harmonic mean of the triangle's sides...Therefore, I think {\bf the answer is 80. But need to verify}...Therefore, the side length of the hexagon is $\boxed{80}$.}\\
    \bottomrule
    \end{tabular}}
    \label{tab:case}
\end{table*}

\section{Related Work}

\paragraph{Large Reasoning Models}
Large Reasoning Models (LRMs) significantly enhances the reasoning capabilities of large language models (LLMs) \citep{o1,xia2025generative}. 
Prominent implementations include OpenAI's o1~\citep{o1}, QwQ \citep{QwQ}, Qwen3 \citep{qwen3}, DeepSeek-R1 \citep{guo2025deepseek}, Claude 3.7 \citep{Claude} and Kimi-1.5 \citep{Kimi-1.5} achieve human-like reasoning by leveraging scaled test-time computation.
In particular, the open-source DeepSeek-R1 utilizes a Mixture-of-Experts (MoE) architecture \citep{DBLP:conf/acl/DaiDZXGCLZYWXLH24} with sparsely activated parameters, selectively activating only 8 out of 256 experts per layer \citep{DBLP:journals/corr/abs-2412-19437}.
This MoE architecture has been widely adopted in recent LLMs \citep{Llama4,DBLP:journals/corr/abs-2409-02060,DBLP:conf/iclr/ShazeerMMDLHD17}, achieving an optimal balance between computational efficiency and competitive performance in complex reasoning tasks.

\paragraph{MoE Models}
Previous research on Mixture of Experts (MoE) models indicates that expert routing is primarily token-dependent \citep{DBLP:conf/icml/XueZFNZZ024}. 
However, recent work \citep{olson2025semantic,dahlke2025mixture} finds that DeepSeek-R1’s advanced reasoning capabilities enable its routing mechanism to achieve greater semantic specialization and structured cognitive processing, representing a substantial advancement over prior MoE models.
Subsequently, \citet{goodfire} train sparse autoencoders (SAEs) on DeepSeek-R1, identifying interpretable features such as backtracking, division, and rapid response patterns within the SAEs space. 
However, training SAEs is computationally intensive, posing significant resource demands.
We employ the normalized Pointwise Mutual Information (nPMI) metric to evaluate expert specialization, requiring only a single forward propagation. 

\paragraph{Efficient Thinking}
Despite significant advancements, large reasoning models continue to encounter substantial cognitive challenges, such as the overthinking \citep{Overthinking,DBLP:journals/corr/abs-2503-16419,DBLP:journals/corr/abs-2502-08235,zaremba2025trading} and underthinking phenomenon \citep{Underthinking,qu2025survey,ballon2025relationship}.
Subsequent efforts address these issues through rule-based stop, decoding constraints \citep{DBLP:journals/corr/abs-2502-10954,Underthinking,DBLP:journals/corr/abs-2502-16235,xiang2025can,ma2025reasoning,DBLP:journals/corr/abs-2501-19393,DBLP:journals/corr/abs-2412-18547,DBLP:journals/corr/abs-2503-05179,zhang2025lightthinker}, steering vectors \citep{chen2025seal,cyberey2025steering}, and parameters tuning \citep{sun2025thinkedit,Overthinking,DBLP:journals/corr/abs-2502-03387,DBLP:journals/corr/abs-2503-04697,DBLP:journals/corr/abs-2412-06769}. 
There are also some works specifically designed to improve reasoning capabilities in MoE architectures by re-mixing experts through gradient-based optimization~\citep{li2025c3po} or by expert pruning via sparse dictionary learning~\citep{tang2025unveiling}.
However, the resource-intensive nature of expert re-mixing algorithms makes them impractical to scale to large models such as 671B-parameter systems, whereas our method is lightweight and directly applicable to such large-scale settings.
Generally, in contrast to the above strategies that primarily rely on crafted rules, extensive labeled data, or computationally expensive parameter training, our \textit{reinforcing cognitive experts} approach achieves more efficient and deeper reasoning with only a single forward propagation, without requiring any supervision signals or additional training.


\section{Conclusion and Future Work}


In this work, we explore ``cognitive experts'' within MoE-based LRMs, proposing a computationally efficient method based on nPMI to identify experts central to cognitive deliberation. We empirically show that steering these experts allows steering of reasoning processes with minimal additional computational overhead or training burden. Critically, the identified experts demonstrated strong transferability across multiple scientific domains, suggesting a fundamental, domain-general cognitive function within the reasoning model. Our findings highlight the promising role of expert specialization as a mechanistic lever for enhancing cognitive control, interpretability, and efficiency in large-scale reasoning architectures.

Future directions include deeper investigations into the structural properties and broader applicability of cognitive experts, as well as integration with other cognitive control strategies to further enhance reasoning robustness. By uncovering this hidden layer of functional specialization within MoE models, we may open new avenues for fine-grained control over neural reasoning processes, more closely mirroring the modularity observed in biological cognitive systems.

itive experts'' within MoE-based LRMs, proposing a computationally efficient method based on nPMI to identify experts central to cognitive deliberation. We empirically show that steering these experts allows precise steering of reasoning processes, effectively resolving inefficiencies like overthinking and underthinking with minimal additional computational overhead or training burden. Critically, the identified experts demonstrated strong transferability across multiple scientific domains, suggesting a fundamental, domain-general cognitive function within the reasoning model. Our findings highlight the promising role of expert specialization as a mechanistic lever for enhancing cognitive control, interpretability, and efficiency in large-scale reasoning architectures.

\section*{Limitations and Broader Impacts}
The internal coordination mechanisms of long-range reasoning models are inherently complex, and our nPMI-based approach may not fully capture all relevant interactions. Future work should explore more sophisticated metrics for expert identification.
Besides, our validation was constrained by the current availability of open-source MoE architectures designed for long-range reasoning, limited to DeepSeek-R1~\citep{guo2025deepseek} and Qwen3-235B~\citep{qwen3}. Additional testing across more diverse architectures is warranted.
The ability to precisely control reasoning processes in large reasoning models has significant implications for both AI safety and efficiency. Our method's minimal computational overhead makes it particularly promising for real-world applications where resource constraints are critical. The observed cross-domain transferability of cognitive experts suggests exciting possibilities for developing more general and adaptable AI systems.

\bibliography{ref}

\begin{thebibliography}{57}
\providecommand{\natexlab}[1]{#1}
\providecommand{\url}[1]{\texttt{#1}}
\expandafter\ifx\csname urlstyle\endcsname\relax
  \providecommand{\doi}[1]{doi: #1}\else
  \providecommand{\doi}{doi: \begingroup \urlstyle{rm}\Url}\fi

\bibitem[Guo et~al.(2025)Guo, Yang, Zhang, Song, Zhang, Xu, Zhu, Ma, Wang, Bi, et~al.]{guo2025deepseek}
Daya Guo, Dejian Yang, Haowei Zhang, Junxiao Song, Ruoyu Zhang, Runxin Xu, Qihao Zhu, Shirong Ma, Peiyi Wang, Xiao Bi, et~al.
\newblock Deepseek-r1: Incentivizing reasoning capability in llms via reinforcement learning.
\newblock \emph{arXiv preprint arXiv:2501.12948}, 2025.

\bibitem[Team(2025)]{qwen3}
Qwen Team.
\newblock Qwen3: Think deeper, act faster.
\newblock 2025.
\newblock URL \url{https://qwenlm.github.io/zh/blog/qwen3/}.

\bibitem[Chen et~al.(2024)Chen, Xu, Liang, He, Pang, Yu, Song, Liu, Zhou, Zhang, Wang, Tu, Mi, and Yu]{Overthinking}
Xingyu Chen, Jiahao Xu, Tian Liang, Zhiwei He, Jianhui Pang, Dian Yu, Linfeng Song, Qiuzhi Liu, Mengfei Zhou, Zhuosheng Zhang, Rui Wang, Zhaopeng Tu, Haitao Mi, and Dong Yu.
\newblock Do {NOT} think that much for 2+3=? on the overthinking of o1-like llms.
\newblock \emph{CoRR}, abs/2412.21187, 2024.
\newblock \doi{10.48550/ARXIV.2412.21187}.
\newblock URL \url{https://doi.org/10.48550/arXiv.2412.21187}.

\bibitem[Wang et~al.(2025{\natexlab{a}})Wang, Liu, Xu, Liang, Chen, He, Song, Yu, Li, Zhang, Wang, Tu, Mi, and Yu]{Underthinking}
Yue Wang, Qiuzhi Liu, Jiahao Xu, Tian Liang, Xingyu Chen, Zhiwei He, Linfeng Song, Dian Yu, Juntao Li, Zhuosheng Zhang, Rui Wang, Zhaopeng Tu, Haitao Mi, and Dong Yu.
\newblock Thoughts are all over the place: On the underthinking of o1-like llms.
\newblock \emph{CoRR}, abs/2501.18585, 2025{\natexlab{a}}.
\newblock \doi{10.48550/ARXIV.2501.18585}.
\newblock URL \url{https://doi.org/10.48550/arXiv.2501.18585}.

\bibitem[Jaech et~al.(2024)Jaech, Kalai, Lerer, Richardson, El-Kishky, Low, Helyar, Madry, Beutel, Carney, et~al.]{o1}
Aaron Jaech, Adam Kalai, Adam Lerer, Adam Richardson, Ahmed El-Kishky, Aiden Low, Alec Helyar, Aleksander Madry, Alex Beutel, Alex Carney, et~al.
\newblock Openai o1 system card.
\newblock \emph{arXiv preprint arXiv:2412.16720}, 2024.

\bibitem[Ji et~al.(2025)Ji, Li, Ye, Wu, Xu, Mo, and Zhang]{ji2025test}
Yixin Ji, Juntao Li, Hai Ye, Kaixin Wu, Jia Xu, Linjian Mo, and Min Zhang.
\newblock Test-time computing: from system-1 thinking to system-2 thinking.
\newblock \emph{arXiv preprint arXiv:2501.02497}, 2025.

\bibitem[Zhang et~al.(2025{\natexlab{a}})Zhang, Lyu, Sun, Wang, Zhang, Guo, Wang, King, Liu, and Ma]{zhang2025and}
Qiyuan Zhang, Fuyuan Lyu, Zexu Sun, Lei Wang, Weixu Zhang, Zhihan Guo, Yufei Wang, Irwin King, Xue Liu, and Chen Ma.
\newblock What, how, where, and how well? a survey on test-time scaling in large language models.
\newblock \emph{arXiv preprint arXiv:2503.24235}, 2025{\natexlab{a}}.

\bibitem[Sui et~al.(2025{\natexlab{a}})Sui, Chuang, Wang, Zhang, Zhang, Yuan, Liu, Wen, Zhong, Chen, et~al.]{sui2025stop}
Yang Sui, Yu-Neng Chuang, Guanchu Wang, Jiamu Zhang, Tianyi Zhang, Jiayi Yuan, Hongyi Liu, Andrew Wen, Shaochen Zhong, Hanjie Chen, et~al.
\newblock Stop overthinking: A survey on efficient reasoning for large language models.
\newblock \emph{arXiv preprint arXiv:2503.16419}, 2025{\natexlab{a}}.

\bibitem[Feng et~al.(2025)Feng, Fang, Ma, and Wang]{feng2025efficient}
Sicheng Feng, Gongfan Fang, Xinyin Ma, and Xinchao Wang.
\newblock Efficient reasoning models: A survey.
\newblock \emph{arXiv preprint arXiv:2504.10903}, 2025.

\bibitem[Pan et~al.(2025)Pan, Ji, Ding, Li, Chen, Wang, Zhou, Chen, Zhang, Wu, et~al.]{pan2025survey}
Qianjun Pan, Wenkai Ji, Yuyang Ding, Junsong Li, Shilian Chen, Junyi Wang, Jie Zhou, Qin Chen, Min Zhang, Yulan Wu, et~al.
\newblock A survey of slow thinking-based reasoning llms using reinforced learning and inference-time scaling law.
\newblock \emph{arXiv preprint arXiv:2505.02665}, 2025.

\bibitem[Qu et~al.(2025)Qu, Li, Su, Sun, Yan, Liu, Cui, Liu, Liang, He, et~al.]{qu2025survey}
Xiaoye Qu, Yafu Li, Zhaochen Su, Weigao Sun, Jianhao Yan, Dongrui Liu, Ganqu Cui, Daizong Liu, Shuxian Liang, Junxian He, et~al.
\newblock A survey of efficient reasoning for large reasoning models: Language, multimodality, and beyond.
\newblock \emph{arXiv preprint arXiv:2503.21614}, 2025.

\bibitem[Chen et~al.(2025{\natexlab{a}})Chen, Qin, Liu, Peng, Guan, Wang, Hu, Zhou, Gao, and Che]{chen2025towards}
Qiguang Chen, Libo Qin, Jinhao Liu, Dengyun Peng, Jiannan Guan, Peng Wang, Mengkang Hu, Yuhang Zhou, Te~Gao, and Wanxiang Che.
\newblock Towards reasoning era: A survey of long chain-of-thought for reasoning large language models.
\newblock \emph{arXiv preprint arXiv:2503.09567}, 2025{\natexlab{a}}.

\bibitem[Wang et~al.(2025{\natexlab{b}})Wang, Wang, Xue, Pang, Liu, Chen, Qiu, Wong, Ji, and Wong]{wang2025harnessing}
Rui Wang, Hongru Wang, Boyang Xue, Jianhui Pang, Shudong Liu, Yi~Chen, Jiahao Qiu, Derek~Fai Wong, Heng Ji, and Kam-Fai Wong.
\newblock Harnessing the reasoning economy: A survey of efficient reasoning for large language models.
\newblock \emph{arXiv preprint arXiv:2503.24377}, 2025{\natexlab{b}}.

\bibitem[Wu et~al.(2025)Wu, Xiang, Wang, and Mittal]{wu2025effectively}
Tong Wu, Chong Xiang, Jiachen~T Wang, and Prateek Mittal.
\newblock Effectively controlling reasoning models through thinking intervention.
\newblock \emph{arXiv preprint arXiv:2503.24370}, 2025.

\bibitem[Lu et~al.(2025)Lu, Han, Acuna, Kim, Jung, Prabhumoye, Muennighoff, Patwary, Shoeybi, Catanzaro, et~al.]{lu2025retro}
Ximing Lu, Seungju Han, David Acuna, Hyunwoo Kim, Jaehun Jung, Shrimai Prabhumoye, Niklas Muennighoff, Mostofa Patwary, Mohammad Shoeybi, Bryan Catanzaro, et~al.
\newblock Retro-search: Exploring untaken paths for deeper and efficient reasoning.
\newblock \emph{arXiv preprint arXiv:2504.04383}, 2025.

\bibitem[Dai et~al.(2024{\natexlab{a}})Dai, Deng, Zhao, Xu, Gao, Chen, Li, Zeng, Yu, Wu, et~al.]{dai2024deepseekmoe}
Damai Dai, Chengqi Deng, Chenggang Zhao, RX~Xu, Huazuo Gao, Deli Chen, Jiashi Li, Wangding Zeng, Xingkai Yu, Yu~Wu, et~al.
\newblock Deepseekmoe: Towards ultimate expert specialization in mixture-of-experts language models.
\newblock \emph{arXiv preprint arXiv:2401.06066}, 2024{\natexlab{a}}.

\bibitem[DeepSeek{-}AI et~al.(2024)DeepSeek{-}AI, Liu, Feng, Xue, Wang, Wu, Lu, Zhao, Deng, Zhang, Ruan, Dai, Guo, Yang, Chen, Ji, Li, Lin, Dai, Luo, Hao, Chen, Li, Zhang, Bao, Xu, Wang, Zhang, Ding, Xin, Gao, Li, Qu, Cai, Liang, Guo, Ni, Li, Wang, Chen, Chen, Yuan, Qiu, Li, Song, Dong, Hu, Gao, Guan, Huang, Yu, Wang, Zhang, Xu, Xia, Zhao, Wang, Zhang, Li, Wang, Zhang, Zhang, Tang, Li, Tian, Huang, Wang, Zhang, Wang, Zhu, Chen, Du, Chen, Jin, Ge, Zhang, Pan, Wang, Xu, Zhang, Chen, Li, Lu, Zhou, Chen, Wu, Ye, Ye, Ma, Wang, Zhou, Yu, Zhou, Pan, Wang, Yun, Pei, Sun, Xiao, and Zeng]{DBLP:journals/corr/abs-2412-19437}
DeepSeek{-}AI, Aixin Liu, Bei Feng, Bing Xue, Bingxuan Wang, Bochao Wu, Chengda Lu, Chenggang Zhao, Chengqi Deng, Chenyu Zhang, Chong Ruan, Damai Dai, Daya Guo, Dejian Yang, Deli Chen, Dongjie Ji, Erhang Li, Fangyun Lin, Fucong Dai, Fuli Luo, Guangbo Hao, Guanting Chen, Guowei Li, H.~Zhang, Han Bao, Hanwei Xu, Haocheng Wang, Haowei Zhang, Honghui Ding, Huajian Xin, Huazuo Gao, Hui Li, Hui Qu, J.~L. Cai, Jian Liang, Jianzhong Guo, Jiaqi Ni, Jiashi Li, Jiawei Wang, Jin Chen, Jingchang Chen, Jingyang Yuan, Junjie Qiu, Junlong Li, Junxiao Song, Kai Dong, Kai Hu, Kaige Gao, Kang Guan, Kexin Huang, Kuai Yu, Lean Wang, Lecong Zhang, Lei Xu, Leyi Xia, Liang Zhao, Litong Wang, Liyue Zhang, Meng Li, Miaojun Wang, Mingchuan Zhang, Minghua Zhang, Minghui Tang, Mingming Li, Ning Tian, Panpan Huang, Peiyi Wang, Peng Zhang, Qiancheng Wang, Qihao Zhu, Qinyu Chen, Qiushi Du, R.~J. Chen, R.~L. Jin, Ruiqi Ge, Ruisong Zhang, Ruizhe Pan, Runji Wang, Runxin Xu, Ruoyu Zhang, Ruyi Chen, S.~S. Li, Shanghao Lu, Shangyan Zhou,
  Shanhuang Chen, Shaoqing Wu, Shengfeng Ye, Shengfeng Ye, Shirong Ma, Shiyu Wang, Shuang Zhou, Shuiping Yu, Shunfeng Zhou, Shuting Pan, T.~Wang, Tao Yun, Tian Pei, Tianyu Sun, W.~L. Xiao, and Wangding Zeng.
\newblock Deepseek-v3 technical report.
\newblock \emph{CoRR}, abs/2412.19437, 2024.
\newblock URL \url{https://doi.org/10.48550/arXiv.2412.19437}.

\bibitem[Xue et~al.(2024)Xue, Zheng, Fu, Ni, Zheng, Zhou, and You]{DBLP:conf/icml/XueZFNZZ024}
Fuzhao Xue, Zian Zheng, Yao Fu, Jinjie Ni, Zangwei Zheng, Wangchunshu Zhou, and Yang You.
\newblock Openmoe: An early effort on open mixture-of-experts language models.
\newblock In \emph{Forty-first International Conference on Machine Learning, {ICML} 2024, Vienna, Austria, July 21-27, 2024}. OpenReview.net, 2024.
\newblock URL \url{https://openreview.net/forum?id=1YDeZU8Lt5}.

\bibitem[Reinhart and Nguyen(2019)]{reinhart2019working}
Robert~MG Reinhart and John~A Nguyen.
\newblock Working memory revived in older adults by synchronizing rhythmic brain circuits.
\newblock \emph{Nature neuroscience}, 22\penalty0 (5):\penalty0 820--827, 2019.

\bibitem[Wischnewski et~al.(2023)Wischnewski, Alekseichuk, and Opitz]{wischnewski2023neurocognitive}
Miles Wischnewski, Ivan Alekseichuk, and Alexander Opitz.
\newblock Neurocognitive, physiological, and biophysical effects of transcranial alternating current stimulation.
\newblock \emph{Trends in Cognitive Sciences}, 27\penalty0 (2):\penalty0 189--205, 2023.

\bibitem[Oathes et~al.(2023)Oathes, Duprat, Reber, Liang, Scully, Long, Deluisi, Sheline, and Linn]{oathes2023non}
Desmond~J Oathes, Romain~JP Duprat, Justin Reber, Ximo Liang, Morgan Scully, Hannah Long, Joseph~A Deluisi, Yvette~I Sheline, and Kristin~A Linn.
\newblock Non-invasively targeting, probing and modulating a deep brain circuit for depression alleviation.
\newblock \emph{Nature Mental Health}, 1\penalty0 (12):\penalty0 1033--1042, 2023.

\bibitem[Grover et~al.(2021)Grover, Nguyen, Viswanathan, and Reinhart]{grover2021high}
Shrey Grover, John~A Nguyen, Vighnesh Viswanathan, and Robert~MG Reinhart.
\newblock High-frequency neuromodulation improves obsessive--compulsive behavior.
\newblock \emph{Nature medicine}, 27\penalty0 (2):\penalty0 232--238, 2021.

\bibitem[Bouma(2009)]{npmi}
Gerlof Bouma.
\newblock Normalized (pointwise) mutual information in collocation extraction.
\newblock \emph{Proceedings of GSCL}, 30:\penalty0 31--40, 2009.

\bibitem[Olson et~al.(2025)Olson, Ratzlaff, Hinck, Luo, Yu, Xue, and Lal]{olson2025semantic}
Matthew~Lyle Olson, Neale Ratzlaff, Musashi Hinck, Man Luo, Sungduk Yu, Chendi Xue, and Vasudev Lal.
\newblock Semantic specialization in moe appears with scale: A study of deepseek r1 expert specialization.
\newblock \emph{arXiv preprint arXiv:2502.10928}, 2025.

\bibitem[\mbox{MAA Committees}()]{aime}
\mbox{MAA Committees}.
\newblock Aime problems and solutions.
\newblock \url{https://artofproblemsolving.com/wiki/index.php/AIME_Problems_and_Solutions}.

\bibitem[Rein et~al.(2024)Rein, Hou, Stickland, Petty, Pang, Dirani, Michael, and Bowman]{gpqa}
David Rein, Betty~Li Hou, Asa~Cooper Stickland, Jackson Petty, Richard~Yuanzhe Pang, Julien Dirani, Julian Michael, and Samuel~R. Bowman.
\newblock {GPQA}: A graduate-level google-proof q\&a benchmark.
\newblock In \emph{First Conference on Language Modeling}, 2024.
\newblock URL \url{https://openreview.net/forum?id=Ti67584b98}.

\bibitem[Li et~al.(2024)Li, Chiang, Frick, Dunlap, Wu, Zhu, Gonzalez, and Stoica]{DBLP:journals/corr/abs-2406-11939}
Tianle Li, Wei{-}Lin Chiang, Evan Frick, Lisa Dunlap, Tianhao Wu, Banghua Zhu, Joseph~E. Gonzalez, and Ion Stoica.
\newblock From crowdsourced data to high-quality benchmarks: Arena-hard and benchbuilder pipeline.
\newblock \emph{CoRR}, abs/2406.11939, 2024.
\newblock URL \url{https://doi.org/10.48550/arXiv.2406.11939}.

\bibitem[Xia et~al.(2025)Xia, Qin, Li, Ma, Fan, Chern, Zou, Zhou, Hu, Jin, et~al.]{xia2025generative}
Shijie Xia, Yiwei Qin, Xuefeng Li, Yan Ma, Run-Ze Fan, Steffi Chern, Haoyang Zou, Fan Zhou, Xiangkun Hu, Jiahe Jin, et~al.
\newblock Generative ai act ii: Test time scaling drives cognition engineering.
\newblock \emph{arXiv preprint arXiv:2504.13828}, 2025.

\bibitem[Qwen(2024)]{QwQ}
Qwen.
\newblock Qwq: Reflect deeply on the boundaries of the unknown.
\newblock 2024.
\newblock URL \url{https://qwenlm.github.io/blog/qwq-32b-preview/}.

\bibitem[Anthropic(2025)]{Claude}
Anthropic.
\newblock Claude 3.7 sonnet.
\newblock 2025.
\newblock URL \url{https://www.anthropic.com/claude/sonnet}.

\bibitem[Team et~al.(2025)Team, Du, Gao, Xing, Jiang, Chen, Li, Xiao, Du, Liao, et~al.]{Kimi-1.5}
Kimi Team, Angang Du, Bofei Gao, Bowei Xing, Changjiu Jiang, Cheng Chen, Cheng Li, Chenjun Xiao, Chenzhuang Du, Chonghua Liao, et~al.
\newblock Kimi k1. 5: Scaling reinforcement learning with llms.
\newblock \emph{arXiv preprint arXiv:2501.12599}, 2025.

\bibitem[Dai et~al.(2024{\natexlab{b}})Dai, Deng, Zhao, Xu, Gao, Chen, Li, Zeng, Yu, Wu, Xie, Li, Huang, Luo, Ruan, Sui, and Liang]{DBLP:conf/acl/DaiDZXGCLZYWXLH24}
Damai Dai, Chengqi Deng, Chenggang Zhao, R.~X. Xu, Huazuo Gao, Deli Chen, Jiashi Li, Wangding Zeng, Xingkai Yu, Y.~Wu, Zhenda Xie, Y.~K. Li, Panpan Huang, Fuli Luo, Chong Ruan, Zhifang Sui, and Wenfeng Liang.
\newblock Deepseekmoe: Towards ultimate expert specialization in mixture-of-experts language models.
\newblock In \emph{Proceedings of the 62nd Annual Meeting of the Association for Computational Linguistics (Volume 1: Long Papers), {ACL} 2024, Bangkok, Thailand, August 11-16, 2024}, pages 1280--1297. Association for Computational Linguistics, 2024{\natexlab{b}}.
\newblock URL \url{https://doi.org/10.18653/v1/2024.acl-long.70}.

\bibitem[Llama(2025)]{Llama4}
Llama.
\newblock The llama 4 herd: The beginning of a new era of natively multimodal ai innovation.
\newblock 2025.
\newblock URL \url{https://www.llama.com/models/llama-4/}.

\bibitem[Muennighoff et~al.(2024)Muennighoff, Soldaini, Groeneveld, Lo, Morrison, Min, Shi, Walsh, Tafjord, Lambert, Gu, Arora, Bhagia, Schwenk, Wadden, Wettig, Hui, Dettmers, Kiela, Farhadi, Smith, Koh, Singh, and Hajishirzi]{DBLP:journals/corr/abs-2409-02060}
Niklas Muennighoff, Luca Soldaini, Dirk Groeneveld, Kyle Lo, Jacob Morrison, Sewon Min, Weijia Shi, Pete Walsh, Oyvind Tafjord, Nathan Lambert, Yuling Gu, Shane Arora, Akshita Bhagia, Dustin Schwenk, David Wadden, Alexander Wettig, Binyuan Hui, Tim Dettmers, Douwe Kiela, Ali Farhadi, Noah~A. Smith, Pang~Wei Koh, Amanpreet Singh, and Hannaneh Hajishirzi.
\newblock Olmoe: Open mixture-of-experts language models.
\newblock \emph{CoRR}, abs/2409.02060, 2024.
\newblock \doi{10.48550/ARXIV.2409.02060}.
\newblock URL \url{https://doi.org/10.48550/arXiv.2409.02060}.

\bibitem[Shazeer et~al.(2017)Shazeer, Mirhoseini, Maziarz, Davis, Le, Hinton, and Dean]{DBLP:conf/iclr/ShazeerMMDLHD17}
Noam Shazeer, Azalia Mirhoseini, Krzysztof Maziarz, Andy Davis, Quoc~V. Le, Geoffrey~E. Hinton, and Jeff Dean.
\newblock Outrageously large neural networks: The sparsely-gated mixture-of-experts layer.
\newblock In \emph{5th International Conference on Learning Representations, {ICLR} 2017, Toulon, France, April 24-26, 2017, Conference Track Proceedings}. OpenReview.net, 2017.
\newblock URL \url{https://openreview.net/forum?id=B1ckMDqlg}.

\bibitem[Dahlke et~al.(2025)Dahlke, Klagges, Zecha, Merkel, Rohr, and Klemm]{dahlke2025mixture}
Robert Dahlke, Henrik Klagges, Dan Zecha, Benjamin Merkel, Sven Rohr, and Fabian Klemm.
\newblock Mixture of tunable experts--behavior modification of deepseek-r1 at inference time.
\newblock \emph{arXiv preprint arXiv:2502.11096}, 2025.

\bibitem[Hazra et~al.(2025)Hazra, Loeffler, Cubuktepe, Avagyan, Gorton, Bissell, Lewis, McGrath, and Balsam]{goodfire}
Dron Hazra, Max Loeffler, Murat Cubuktepe, Levon Avagyan, Liv Gorton, Mark Bissell, Owen Lewis, Thomas McGrath, and Daniel Balsam.
\newblock Under the hood of a reasoning model.
\newblock 2025.
\newblock URL \url{https://www.goodfire.ai/blog/under-the-hood-of-a-reasoning-model}.

\bibitem[Sui et~al.(2025{\natexlab{b}})Sui, Chuang, Wang, Zhang, Zhang, Yuan, Liu, Wen, Zhong, Chen, and Hu]{DBLP:journals/corr/abs-2503-16419}
Yang Sui, Yu{-}Neng Chuang, Guanchu Wang, Jiamu Zhang, Tianyi Zhang, Jiayi Yuan, Hongyi Liu, Andrew Wen, Shaochen Zhong, Hanjie Chen, and Xia~Ben Hu.
\newblock Stop overthinking: {A} survey on efficient reasoning for large language models.
\newblock \emph{CoRR}, abs/2503.16419, 2025{\natexlab{b}}.
\newblock \doi{10.48550/ARXIV.2503.16419}.
\newblock URL \url{https://doi.org/10.48550/arXiv.2503.16419}.

\bibitem[Cuadron et~al.(2025)Cuadron, Li, Ma, Wang, Wang, Zhuang, Liu, Schroeder, Xia, Mao, Thumiger, Desai, Stoica, Klimovic, Neubig, and Gonzalez]{DBLP:journals/corr/abs-2502-08235}
Alejandro Cuadron, Dacheng Li, Wenjie Ma, Xingyao Wang, Yichuan Wang, Siyuan Zhuang, Shu Liu, Luis~Gaspar Schroeder, Tian Xia, Huanzhi Mao, Nicholas Thumiger, Aditya Desai, Ion Stoica, Ana Klimovic, Graham Neubig, and Joseph~E. Gonzalez.
\newblock The danger of overthinking: Examining the reasoning-action dilemma in agentic tasks.
\newblock \emph{CoRR}, abs/2502.08235, 2025.
\newblock \doi{10.48550/ARXIV.2502.08235}.
\newblock URL \url{https://doi.org/10.48550/arXiv.2502.08235}.

\bibitem[Zaremba et~al.(2025)Zaremba, Nitishinskaya, Barak, Lin, Toyer, Yu, Dias, Wallace, Xiao, Heidecke, et~al.]{zaremba2025trading}
Wojciech Zaremba, Evgenia Nitishinskaya, Boaz Barak, Stephanie Lin, Sam Toyer, Yaodong Yu, Rachel Dias, Eric Wallace, Kai Xiao, Johannes Heidecke, et~al.
\newblock Trading inference-time compute for adversarial robustness.
\newblock \emph{arXiv preprint arXiv:2501.18841}, 2025.

\bibitem[Ballon et~al.(2025)Ballon, Algaba, and Ginis]{ballon2025relationship}
Marthe Ballon, Andres Algaba, and Vincent Ginis.
\newblock The relationship between reasoning and performance in large language models--o3 (mini) thinks harder, not longer.
\newblock \emph{arXiv preprint arXiv:2502.15631}, 2025.

\bibitem[Tran et~al.(2025)Tran, Dat, Anh, and Thanh{-}Tung]{DBLP:journals/corr/abs-2502-10954}
Bao~Hieu Tran, Nguyen~Cong Dat, Nguyen~Duc Anh, and Hoang Thanh{-}Tung.
\newblock Learning to stop overthinking at test time.
\newblock \emph{CoRR}, abs/2502.10954, 2025.
\newblock \doi{10.48550/ARXIV.2502.10954}.
\newblock URL \url{https://doi.org/10.48550/arXiv.2502.10954}.

\bibitem[Ding et~al.(2025)Ding, Jiang, Liu, Jing, Guo, Wang, Zhang, Wang, Liu, Du, Liu, and Tao]{DBLP:journals/corr/abs-2502-16235}
Yifu Ding, Wentao Jiang, Shunyu Liu, Yongcheng Jing, Jinyang Guo, Yingjie Wang, Jing Zhang, Zengmao Wang, Ziwei Liu, Bo~Du, Xianglong Liu, and Dacheng Tao.
\newblock Dynamic parallel tree search for efficient {LLM} reasoning.
\newblock \emph{CoRR}, abs/2502.16235, 2025.
\newblock \doi{10.48550/ARXIV.2502.16235}.
\newblock URL \url{https://doi.org/10.48550/arXiv.2502.16235}.

\bibitem[Xiang et~al.(2025)Xiang, Liu, Jiang, Nie, Cai, Yin, Huang, Fan, Li, Huang, et~al.]{xiang2025can}
Kun Xiang, Zhili Liu, Zihao Jiang, Yunshuang Nie, Kaixin Cai, Yiyang Yin, Runhui Huang, Haoxiang Fan, Hanhui Li, Weiran Huang, et~al.
\newblock Can atomic step decomposition enhance the self-structured reasoning of multimodal large models?
\newblock \emph{arXiv preprint arXiv:2503.06252}, 2025.

\bibitem[Ma et~al.(2025)Ma, He, Snell, Griggs, Min, and Zaharia]{ma2025reasoning}
Wenjie Ma, Jingxuan He, Charlie Snell, Tyler Griggs, Sewon Min, and Matei Zaharia.
\newblock Reasoning models can be effective without thinking.
\newblock \emph{arXiv preprint arXiv:2504.09858}, 2025.

\bibitem[Muennighoff et~al.(2025)Muennighoff, Yang, Shi, Li, Fei{-}Fei, Hajishirzi, Zettlemoyer, Liang, Cand{\`{e}}s, and Hashimoto]{DBLP:journals/corr/abs-2501-19393}
Niklas Muennighoff, Zitong Yang, Weijia Shi, Xiang~Lisa Li, Li~Fei{-}Fei, Hannaneh Hajishirzi, Luke Zettlemoyer, Percy Liang, Emmanuel~J. Cand{\`{e}}s, and Tatsunori Hashimoto.
\newblock s1: Simple test-time scaling.
\newblock \emph{CoRR}, abs/2501.19393, 2025.
\newblock \doi{10.48550/ARXIV.2501.19393}.
\newblock URL \url{https://doi.org/10.48550/arXiv.2501.19393}.

\bibitem[Han et~al.(2024)Han, Wang, Fang, Zhao, Ma, and Chen]{DBLP:journals/corr/abs-2412-18547}
Tingxu Han, Zhenting Wang, Chunrong Fang, Shiyu Zhao, Shiqing Ma, and Zhenyu Chen.
\newblock Token-budget-aware {LLM} reasoning.
\newblock \emph{CoRR}, abs/2412.18547, 2024.
\newblock \doi{10.48550/ARXIV.2412.18547}.
\newblock URL \url{https://doi.org/10.48550/arXiv.2412.18547}.

\bibitem[Aytes et~al.(2025)Aytes, Baek, and Hwang]{DBLP:journals/corr/abs-2503-05179}
Simon~A. Aytes, Jinheon Baek, and Sung~Ju Hwang.
\newblock Sketch-of-thought: Efficient {LLM} reasoning with adaptive cognitive-inspired sketching.
\newblock \emph{CoRR}, abs/2503.05179, 2025.
\newblock \doi{10.48550/ARXIV.2503.05179}.
\newblock URL \url{https://doi.org/10.48550/arXiv.2503.05179}.

\bibitem[Zhang et~al.(2025{\natexlab{b}})Zhang, Zhu, Sun, Luo, Qiao, Du, Zheng, Chen, and Zhang]{zhang2025lightthinker}
Jintian Zhang, Yuqi Zhu, Mengshu Sun, Yujie Luo, Shuofei Qiao, Lun Du, Da~Zheng, Huajun Chen, and Ningyu Zhang.
\newblock Lightthinker: Thinking step-by-step compression.
\newblock \emph{arXiv preprint arXiv:2502.15589}, 2025{\natexlab{b}}.

\bibitem[Chen et~al.(2025{\natexlab{b}})Chen, Zhang, Hong, Kundu, and Wang]{chen2025seal}
Runjin Chen, Zhenyu Zhang, Junyuan Hong, Souvik Kundu, and Zhangyang Wang.
\newblock Seal: Steerable reasoning calibration of large language models for free.
\newblock \emph{arXiv preprint arXiv:2504.07986}, 2025{\natexlab{b}}.

\bibitem[Cyberey and Evans(2025)]{cyberey2025steering}
Hannah Cyberey and David Evans.
\newblock Steering the censorship: Uncovering representation vectors for llm" thought" control.
\newblock \emph{arXiv preprint arXiv:2504.17130}, 2025.

\bibitem[Sun et~al.(2025)Sun, Yan, and Weng]{sun2025thinkedit}
Chung-En Sun, Ge~Yan, and Tsui-Wei Weng.
\newblock Thinkedit: Interpretable weight editing to mitigate overly short thinking in reasoning models.
\newblock \emph{arXiv preprint arXiv:2503.22048}, 2025.

\bibitem[Ye et~al.(2025)Ye, Huang, Xiao, Chern, Xia, and Liu]{DBLP:journals/corr/abs-2502-03387}
Yixin Ye, Zhen Huang, Yang Xiao, Ethan Chern, Shijie Xia, and Pengfei Liu.
\newblock {LIMO:} less is more for reasoning.
\newblock \emph{CoRR}, abs/2502.03387, 2025.
\newblock \doi{10.48550/ARXIV.2502.03387}.
\newblock URL \url{https://doi.org/10.48550/arXiv.2502.03387}.

\bibitem[Aggarwal and Welleck(2025)]{DBLP:journals/corr/abs-2503-04697}
Pranjal Aggarwal and Sean Welleck.
\newblock {L1:} controlling how long {A} reasoning model thinks with reinforcement learning.
\newblock \emph{CoRR}, abs/2503.04697, 2025.
\newblock \doi{10.48550/ARXIV.2503.04697}.
\newblock URL \url{https://doi.org/10.48550/arXiv.2503.04697}.

\bibitem[Hao et~al.(2024)Hao, Sukhbaatar, Su, Li, Hu, Weston, and Tian]{DBLP:journals/corr/abs-2412-06769}
Shibo Hao, Sainbayar Sukhbaatar, DiJia Su, Xian Li, Zhiting Hu, Jason Weston, and Yuandong Tian.
\newblock Training large language models to reason in a continuous latent space.
\newblock \emph{CoRR}, abs/2412.06769, 2024.
\newblock \doi{10.48550/ARXIV.2412.06769}.
\newblock URL \url{https://doi.org/10.48550/arXiv.2412.06769}.

\bibitem[Li et~al.(2025)Li, Li, and Zhou]{li2025c3po}
Zhongyang Li, Ziyue Li, and Tianyi Zhou.
\newblock C3po: Critical-layer, core-expert, collaborative pathway optimization for test-time expert re-mixing.
\newblock \emph{arXiv preprint arXiv:2504.07964}, 2025.

\bibitem[Tang et~al.(2025)Tang, Tang, Zhang, Chen, and Li]{tang2025unveiling}
Yuanbo Tang, Yan Tang, Naifan Zhang, Meixuan Chen, and Yang Li.
\newblock Unveiling hidden collaboration within mixture-of-experts in large language models.
\newblock \emph{arXiv preprint arXiv:2504.12359}, 2025.

\end{thebibliography}


\begin{thebibliography}{44}
\providecommand{\natexlab}[1]{#1}
\providecommand{\url}[1]{\texttt{#1}}
\expandafter\ifx\csname urlstyle\endcsname\relax
  \providecommand{\doi}[1]{doi: #1}\else
  \providecommand{\doi}{doi: \begingroup \urlstyle{rm}\Url}\fi

\bibitem[qwe(2024)]{qwen2}
Qwen2 technical report.
\newblock 2024.

\bibitem[Abdin et~al.(2024)Abdin, Jacobs, Awan, Aneja, Awadallah, Awadalla, Bach, Bahree, Bakhtiari, Behl, et~al.]{abdin2024phi}
Marah Abdin, Sam~Ade Jacobs, Ammar~Ahmad Awan, Jyoti Aneja, Ahmed Awadallah, Hany Awadalla, Nguyen Bach, Amit Bahree, Arash Bakhtiari, Harkirat Behl, et~al.
\newblock Phi-3 technical report: A highly capable language model locally on your phone.
\newblock \emph{arXiv preprint arXiv:2404.14219}, 2024.

\bibitem[Achiam et~al.(2023)Achiam, Adler, Agarwal, Ahmad, Akkaya, Aleman, Almeida, Altenschmidt, Altman, Anadkat, et~al.]{achiam2023gpt}
Josh Achiam, Steven Adler, Sandhini Agarwal, Lama Ahmad, Ilge Akkaya, Florencia~Leoni Aleman, Diogo Almeida, Janko Altenschmidt, Sam Altman, Shyamal Anadkat, et~al.
\newblock Gpt-4 technical report.
\newblock \emph{arXiv preprint arXiv:2303.08774}, 2023.

\bibitem[Bai et~al.(2024)Bai, Du, Liang, Jin, Liu, Zhou, Zheng, Zhang, Ma, Wang, et~al.]{bai2024coig}
Yuelin Bai, Xinrun Du, Yiming Liang, Yonggang Jin, Ziqiang Liu, Junting Zhou, Tianyu Zheng, Xincheng Zhang, Nuo Ma, Zekun Wang, et~al.
\newblock Coig-cqia: Quality is all you need for chinese instruction fine-tuning.
\newblock \emph{arXiv preprint arXiv:2403.18058}, 2024.

\bibitem[Bauer et~al.(2024)Bauer, Trapp, Stenger, Leppich, Kounev, Leznik, Chard, and Foster]{bauer2024comprehensive}
Andr{\'e} Bauer, Simon Trapp, Michael Stenger, Robert Leppich, Samuel Kounev, Mark Leznik, Kyle Chard, and Ian Foster.
\newblock Comprehensive exploration of synthetic data generation: A survey.
\newblock \emph{arXiv preprint arXiv:2401.02524}, 2024.

\bibitem[Bi et~al.(2024)Bi, Chen, Chen, Chen, Dai, Deng, Ding, Dong, Du, Fu, et~al.]{bi2024deepseek}
Xiao Bi, Deli Chen, Guanting Chen, Shanhuang Chen, Damai Dai, Chengqi Deng, Honghui Ding, Kai Dong, Qiushi Du, Zhe Fu, et~al.
\newblock Deepseek llm: Scaling open-source language models with longtermism.
\newblock \emph{arXiv preprint arXiv:2401.02954}, 2024.

\bibitem[Broder(1997)]{broder1997resemblance}
Andrei~Z Broder.
\newblock On the resemblance and containment of documents.
\newblock In \emph{Proceedings. Compression and Complexity of SEQUENCES 1997 (Cat. No. 97TB100171)}, pp.\  21--29. IEEE, 1997.

\bibitem[Cai et~al.(2023)Cai, Wang, Ma, Chen, and Zhou]{cai2023large}
Tianle Cai, Xuezhi Wang, Tengyu Ma, Xinyun Chen, and Denny Zhou.
\newblock Large language models as tool makers.
\newblock \emph{arXiv preprint arXiv:2305.17126}, 2023.

\bibitem[Chakraborty et~al.(2023)Chakraborty, Bedi, Zhu, An, Manocha, and Huang]{chakraborty2023possibilities}
Souradip Chakraborty, Amrit~Singh Bedi, Sicheng Zhu, Bang An, Dinesh Manocha, and Furong Huang.
\newblock On the possibilities of ai-generated text detection.
\newblock \emph{arXiv preprint arXiv:2304.04736}, 2023.

\bibitem[Chen et~al.(2022)Chen, Ma, Wang, and Cohen]{chen2022program}
Wenhu Chen, Xueguang Ma, Xinyi Wang, and William~W Cohen.
\newblock Program of thoughts prompting: Disentangling computation from reasoning for numerical reasoning tasks.
\newblock \emph{arXiv preprint arXiv:2211.12588}, 2022.

\bibitem[Choi \& Li(2024)Choi and Li]{choipicle}
Hyeong~Kyu Choi and Yixuan Li.
\newblock Picle: Eliciting diverse behaviors from large language models with persona in-context learning.
\newblock In \emph{Forty-first International Conference on Machine Learning}, 2024.

\bibitem[Computer(2023)]{together2023redpajama}
Together Computer.
\newblock Redpajama: an open dataset for training large language models, 2023.
\newblock URL \url{https://github.com/togethercomputer/RedPajama-Data}.

\bibitem[Del{\'e}tang et~al.(2023)Del{\'e}tang, Ruoss, Duquenne, Catt, Genewein, Mattern, Grau-Moya, Wenliang, Aitchison, Orseau, et~al.]{deletang2023language}
Gr{\'e}goire Del{\'e}tang, Anian Ruoss, Paul-Ambroise Duquenne, Elliot Catt, Tim Genewein, Christopher Mattern, Jordi Grau-Moya, Li~Kevin Wenliang, Matthew Aitchison, Laurent Orseau, et~al.
\newblock Language modeling is compression.
\newblock \emph{arXiv preprint arXiv:2309.10668}, 2023.

\bibitem[Dohmatob et~al.(2024)Dohmatob, Feng, Yang, Charton, and Kempe]{dohmatob2024tale}
Elvis Dohmatob, Yunzhen Feng, Pu~Yang, Francois Charton, and Julia Kempe.
\newblock A tale of tails: Model collapse as a change of scaling laws.
\newblock \emph{arXiv preprint arXiv:2402.07043}, 2024.

\bibitem[Gandhi et~al.(2023)Gandhi, Sadigh, and Goodman]{gandhi2023strategic}
Kanishk Gandhi, Dorsa Sadigh, and Noah~D Goodman.
\newblock Strategic reasoning with language models.
\newblock \emph{arXiv preprint arXiv:2305.19165}, 2023.

\bibitem[Ge et~al.(2024)Ge, Jing, Wang, Wang, Chen, and Wei]{ge2024incontext}
Tao Ge, Hu~Jing, Lei Wang, Xun Wang, Si-Qing Chen, and Furu Wei.
\newblock In-context autoencoder for context compression in a large language model.
\newblock In \emph{The Twelfth International Conference on Learning Representations}, 2024.
\newblock URL \url{https://openreview.net/forum?id=uREj4ZuGJE}.

\bibitem[Hendrycks et~al.(2021)Hendrycks, Burns, Kadavath, Arora, Basart, Tang, Song, and Steinhardt]{hendrycks2021measuring}
Dan Hendrycks, Collin Burns, Saurav Kadavath, Akul Arora, Steven Basart, Eric Tang, Dawn Song, and Jacob Steinhardt.
\newblock Measuring mathematical problem solving with the math dataset.
\newblock \emph{arXiv preprint arXiv:2103.03874}, 2021.

\bibitem[Huang et~al.(2024)Huang, Liu, Gong, Gou, Shen, Duan, and Chen]{huang2024key}
Yiming Huang, Xiao Liu, Yeyun Gong, Zhibin Gou, Yelong Shen, Nan Duan, and Weizhu Chen.
\newblock Key-point-driven data synthesis with its enhancement on mathematical reasoning.
\newblock \emph{arXiv preprint arXiv:2403.02333}, 2024.

\bibitem[Jandaghi et~al.(2023)Jandaghi, Sheng, Bai, Pujara, and Sidahmed]{jandaghi2023faithful}
Pegah Jandaghi, XiangHai Sheng, Xinyi Bai, Jay Pujara, and Hakim Sidahmed.
\newblock Faithful persona-based conversational dataset generation with large language models.
\newblock \emph{arXiv preprint arXiv:2312.10007}, 2023.

\bibitem[Kaplan et~al.(2020)Kaplan, McCandlish, Henighan, Brown, Chess, Child, Gray, Radford, Wu, and Amodei]{kaplan2020scaling}
Jared Kaplan, Sam McCandlish, Tom Henighan, Tom~B Brown, Benjamin Chess, Rewon Child, Scott Gray, Alec Radford, Jeffrey Wu, and Dario Amodei.
\newblock Scaling laws for neural language models.
\newblock \emph{arXiv preprint arXiv:2001.08361}, 2020.

\bibitem[Li et~al.(2024{\natexlab{a}})Li, Wang, Hu, Wei, Zheng, Hu, Zhang, and Peng]{li2024common}
Chen Li, Weiqi Wang, Jingcheng Hu, Yixuan Wei, Nanning Zheng, Han Hu, Zheng Zhang, and Houwen Peng.
\newblock Common 7b language models already possess strong math capabilities.
\newblock \emph{arXiv preprint arXiv:2403.04706}, 2024{\natexlab{a}}.

\bibitem[Li et~al.(2024{\natexlab{b}})Li, Dong, Tang, Wang, Zhang, Huang, Huang, Huang, Huang, Zhang, et~al.]{li2024synthetic}
Haoran Li, Qingxiu Dong, Zhengyang Tang, Chaojun Wang, Xingxing Zhang, Haoyang Huang, Shaohan Huang, Xiaolong Huang, Zeqiang Huang, Dongdong Zhang, et~al.
\newblock Synthetic data (almost) from scratch: Generalized instruction tuning for language models.
\newblock \emph{arXiv preprint arXiv:2402.13064}, 2024{\natexlab{b}}.

\bibitem[Li et~al.(2023{\natexlab{a}})Li, Mehrabi, Peris, Goyal, Chang, Galstyan, Zemel, and Gupta]{li2023steerability}
Junyi Li, Ninareh Mehrabi, Charith Peris, Palash Goyal, Kai-Wei Chang, Aram Galstyan, Richard Zemel, and Rahul Gupta.
\newblock On the steerability of large language models toward data-driven personas.
\newblock \emph{arXiv preprint arXiv:2311.04978}, 2023{\natexlab{a}}.

\bibitem[Li et~al.(2023{\natexlab{b}})Li, Bubeck, Eldan, Del~Giorno, Gunasekar, and Lee]{li2023textbooks}
Yuanzhi Li, S{\'e}bastien Bubeck, Ronen Eldan, Allie Del~Giorno, Suriya Gunasekar, and Yin~Tat Lee.
\newblock Textbooks are all you need ii: phi-1.5 technical report.
\newblock \emph{arXiv preprint arXiv:2309.05463}, 2023{\natexlab{b}}.

\bibitem[Liu et~al.(2024)Liu, Wei, Liu, Si, Zhang, Rao, Zheng, Peng, Yang, Zhou, et~al.]{liu2024best}
Ruibo Liu, Jerry Wei, Fangyu Liu, Chenglei Si, Yanzhe Zhang, Jinmeng Rao, Steven Zheng, Daiyi Peng, Diyi Yang, Denny Zhou, et~al.
\newblock Best practices and lessons learned on synthetic data for language models.
\newblock \emph{arXiv preprint arXiv:2404.07503}, 2024.

\bibitem[Liu et~al.(2023)Liu, Zhang, Li, Liu, and Yang]{liu2023dynamic}
Zijun Liu, Yanzhe Zhang, Peng Li, Yang Liu, and Diyi Yang.
\newblock Dynamic llm-agent network: An llm-agent collaboration framework with agent team optimization.
\newblock \emph{arXiv preprint arXiv:2310.02170}, 2023.

\bibitem[Maini et~al.(2024)Maini, Seto, Bai, Grangier, Zhang, and Jaitly]{maini2024rephrasing}
Pratyush Maini, Skyler Seto, He~Bai, David Grangier, Yizhe Zhang, and Navdeep Jaitly.
\newblock Rephrasing the web: A recipe for compute and data-efficient language modeling.
\newblock \emph{arXiv preprint arXiv:2401.16380}, 2024.

\bibitem[Pan et~al.(2023)Pan, Pan, Chen, Nakov, Kan, and Wang]{pan2023risk}
Yikang Pan, Liangming Pan, Wenhu Chen, Preslav Nakov, Min-Yen Kan, and William~Yang Wang.
\newblock On the risk of misinformation pollution with large language models.
\newblock \emph{arXiv preprint arXiv:2305.13661}, 2023.

\bibitem[Park et~al.(2023)Park, O'Brien, Cai, Morris, Liang, and Bernstein]{park2023generative}
Joon~Sung Park, Joseph O'Brien, Carrie~Jun Cai, Meredith~Ringel Morris, Percy Liang, and Michael~S Bernstein.
\newblock Generative agents: Interactive simulacra of human behavior.
\newblock In \emph{Proceedings of the 36th Annual ACM Symposium on User Interface Software and Technology}, pp.\  1--22, 2023.

\bibitem[Schick et~al.(2024)Schick, Dwivedi-Yu, Dess{\`\i}, Raileanu, Lomeli, Hambro, Zettlemoyer, Cancedda, and Scialom]{schick2024toolformer}
Timo Schick, Jane Dwivedi-Yu, Roberto Dess{\`\i}, Roberta Raileanu, Maria Lomeli, Eric Hambro, Luke Zettlemoyer, Nicola Cancedda, and Thomas Scialom.
\newblock Toolformer: Language models can teach themselves to use tools.
\newblock \emph{Advances in Neural Information Processing Systems}, 36, 2024.

\bibitem[Shanahan et~al.(2023)Shanahan, McDonell, and Reynolds]{shanahan2023role}
Murray Shanahan, Kyle McDonell, and Laria Reynolds.
\newblock Role play with large language models.
\newblock \emph{Nature}, 623\penalty0 (7987):\penalty0 493--498, 2023.

\bibitem[Shumailov et~al.(2023)Shumailov, Shumaylov, Zhao, Gal, Papernot, and Anderson]{shumailov2023curse}
Ilia Shumailov, Zakhar Shumaylov, Yiren Zhao, Yarin Gal, Nicolas Papernot, and Ross Anderson.
\newblock The curse of recursion: Training on generated data makes models forget.
\newblock \emph{arXiv preprint arXiv:2305.17493}, 2023.

\bibitem[Team(2024)]{qwen1.5}
Qwen Team.
\newblock Introducing qwen1.5, February 2024.
\newblock URL \url{https://qwenlm.github.io/blog/qwen1.5/}.

\bibitem[Travers \& Milgram(1977)Travers and Milgram]{travers1977experimental}
Jeffrey Travers and Stanley Milgram.
\newblock An experimental study of the small world problem.
\newblock In \emph{Social networks}, pp.\  179--197. Elsevier, 1977.

\bibitem[Villalobos et~al.(2024)Villalobos, Ho, Sevilla, Besiroglu, Heim, and Hobbhahn]{villalobosposition}
Pablo Villalobos, Anson Ho, Jaime Sevilla, Tamay Besiroglu, Lennart Heim, and Marius Hobbhahn.
\newblock Position: Will we run out of data? limits of llm scaling based on human-generated data.
\newblock In \emph{Forty-first International Conference on Machine Learning}, 2024.

\bibitem[Wang et~al.(2023)Wang, Ren, Zhou, Lu, Luo, Shi, Zhang, Song, Zhan, and Li]{wang2023mathcoder}
Ke~Wang, Houxing Ren, Aojun Zhou, Zimu Lu, Sichun Luo, Weikang Shi, Renrui Zhang, Linqi Song, Mingjie Zhan, and Hongsheng Li.
\newblock Mathcoder: Seamless code integration in llms for enhanced mathematical reasoning.
\newblock \emph{arXiv preprint arXiv:2310.03731}, 2023.

\bibitem[Wang et~al.(2022)Wang, Kordi, Mishra, Liu, Smith, Khashabi, and Hajishirzi]{wang2022self}
Yizhong Wang, Yeganeh Kordi, Swaroop Mishra, Alisa Liu, Noah~A Smith, Daniel Khashabi, and Hannaneh Hajishirzi.
\newblock Self-instruct: Aligning language models with self-generated instructions.
\newblock \emph{arXiv preprint arXiv:2212.10560}, 2022.

\bibitem[Wang et~al.(2024)Wang, Mao, Wu, Ge, Wei, and Ji]{wang2024unleashing}
Zhenhailong Wang, Shaoguang Mao, Wenshan Wu, Tao Ge, Furu Wei, and Heng Ji.
\newblock Unleashing the emergent cognitive synergy in large language models: A task-solving agent through multi-persona self-collaboration.
\newblock In \emph{Proceedings of the 2024 Conference of the North American Chapter of the Association for Computational Linguistics: Human Language Technologies (Volume 1: Long Papers)}, pp.\  257--279, 2024.

\bibitem[Xu et~al.(2024)Xu, Jain, and Kankanhalli]{xu2024hallucination}
Ziwei Xu, Sanjay Jain, and Mohan Kankanhalli.
\newblock Hallucination is inevitable: An innate limitation of large language models.
\newblock \emph{arXiv preprint arXiv:2401.11817}, 2024.

\bibitem[Young et~al.(2024)Young, Chen, Li, Huang, Zhang, Zhang, Li, Zhu, Chen, Chang, et~al.]{young2024yi}
Alex Young, Bei Chen, Chao Li, Chengen Huang, Ge~Zhang, Guanwei Zhang, Heng Li, Jiangcheng Zhu, Jianqun Chen, Jing Chang, et~al.
\newblock Yi: Open foundation models by 01. ai.
\newblock \emph{arXiv preprint arXiv:2403.04652}, 2024.

\bibitem[Yu et~al.(2023)Yu, Jiang, Shi, Yu, Liu, Zhang, Kwok, Li, Weller, and Liu]{yu2023metamath}
Longhui Yu, Weisen Jiang, Han Shi, Jincheng Yu, Zhengying Liu, Yu~Zhang, James~T Kwok, Zhenguo Li, Adrian Weller, and Weiyang Liu.
\newblock Metamath: Bootstrap your own mathematical questions for large language models.
\newblock \emph{arXiv preprint arXiv:2309.12284}, 2023.

\bibitem[Zhang et~al.(2024)Zhang, Mao, Ge, Wang, de~Wynter, Xia, Wu, Song, Lan, and Wei]{zhang2024llm}
Yadong Zhang, Shaoguang Mao, Tao Ge, Xun Wang, Adrian de~Wynter, Yan Xia, Wenshan Wu, Ting Song, Man Lan, and Furu Wei.
\newblock Llm as a mastermind: A survey of strategic reasoning with large language models.
\newblock \emph{arXiv preprint arXiv:2404.01230}, 2024.

\bibitem[Zhao et~al.(2024)Zhao, Ren, Hessel, Cardie, Choi, and Deng]{zhao2024wildchat}
Wenting Zhao, Xiang Ren, Jack Hessel, Claire Cardie, Yejin Choi, and Yuntian Deng.
\newblock Wildchat: 1m chat{GPT} interaction logs in the wild.
\newblock In \emph{The Twelfth International Conference on Learning Representations}, 2024.
\newblock URL \url{https://openreview.net/forum?id=Bl8u7ZRlbM}.

\bibitem[Zhu et~al.(2024)Zhu, Guo, Shao, Yang, Wang, Xu, Wu, Li, Gao, Ma, et~al.]{zhu2024deepseek}
Qihao Zhu, Daya Guo, Zhihong Shao, Dejian Yang, Peiyi Wang, Runxin Xu, Y~Wu, Yukun Li, Huazuo Gao, Shirong Ma, et~al.
\newblock Deepseek-coder-v2: Breaking the barrier of closed-source models in code intelligence.
\newblock \emph{arXiv preprint arXiv:2406.11931}, 2024.

\end{thebibliography}
\bibliographystyle{unsrtnat}

\newpage
\appendix

\section{Experiment Setup}
\label{appendix:Experiment}

\subsection{Baselines}
\label{appendix:baselines}

We evaluate our cognitive experts in comparison with two widely used inference-time techniques for reasoning tasks: prompt engineering and decoding constraint. In particular, we consider two types of prompt placements in our analysis — one positioned before the \texttt{<think>} token ($\text{Prompt}{\text{before}}$) and the other placed after it ($\text{Prompt}{\text{after}}$), defined as follows:

\begin{tcolorbox}[title=\textbf{Prompt before} \texttt{<think>},
                  colback=white,
                  colframe=Gray!80,
                  colbacktitle=Gray!5,
                  coltitle=black,
                  rounded corners,
                  boxrule=0.8pt,
                  sharp corners]
$<$\textbar{}begin\_of\_sentence\textbar{}$>$$<$\textbar{}User\textbar{}$>$
$<$context$>$

You are an expert math-solving assistant who prioritizes clear, concise solutions. You solve problems in a single thought process, ensuring accuracy and efficiency. You seek clarification when needed and respect user preferences even if they are unconventional.

$<$/context$>$\\

$<$solving\_rules$>$\\
- Try to complete every idea you think of and don't give up halfway\\
- Don't skip steps\\
- Display solution process clearly\\
- Ask for clarification on ambiguity\\
$<$/solving\_rules$>$\\

$<$format\_rules$>$\\
- Use equations and explanations for clarity\\
- Keep responses brief but complete\\
- Provide step-by-step reasoning if needed\\
$<$/format\_rules$>$\\

PROBLEM: \{problem\}\\

OUTPUT: Please think carefully and follow above rules to get the correct answer for PROBLEM. Focus on clear, concise solutions while maintaining a helpful, accurate style.$<$\textbar{}Assistant\textbar{}$>$\hlthinkblue{think}\textbackslash n
\end{tcolorbox}

\begin{tcolorbox}[title=\textbf{Prompt after} \texttt{<think>},
                  colback=white,
                  colframe=Gray!80,
                  colbacktitle=Gray!5,
                  coltitle=black,
                  rounded corners,
                  boxrule=0.8pt,
                  sharp corners]
$<$\textbar{}begin\_of\_sentence\textbar{}$>$$<$\textbar{}User\textbar{}$>$
$<$context$>$

You are an expert math-solving assistant who prioritizes clear, concise solutions. You solve problems in a single thought process, ensuring accuracy and efficiency. You seek clarification when needed and respect user preferences even if they are unconventional.

$<$/context$>$\\

PROBLEM: \{problem\}\\

\hlthinkblue{think}\textbackslash n \\

Please think carefully and follow these rules to find the correct answer for PROBLEM. \\

$<$solving\_rules$>$\\
- Try to complete every idea you think of and don't give up halfway\\
- Don't skip steps\\
- Display solution process clearly\\
- Ask for clarification on ambiguity\\
$<$/solving\_rules$>$\\

$<$format\_rules$>$\\
- Use equations and explanations for clarity\\
- Keep responses brief but complete\\
- Provide step-by-step reasoning if needed\\
$<$/format\_rules$>$\\

Focus on clear, concise solutions while maintaining a helpful and accurate style.\\

OUTPUT: 
\end{tcolorbox}

\subsection{Experiments Compute Resources}
\label{appendix:gpu}
We conduct our DeepSeek-R1 experiments on 16 H20 GPUs using \texttt{vllm==0.7.0}. It is worth noting that for experiments on the Qwen3-235B-A22B model, we use \texttt{vllm==0.8.5.post} because the recently released Qwen3-235B-A22B models are only compatible with \texttt{vllm} versions \(\geq\) 0.8.5.

\section{Experiment details and Results}

\subsection{Cognitive Experts of Qwen3-235B}
\label{appendix:expert_of_qwen}

We employ Qwen3-235B to generate responses on the AIME2024 dataset, while recording the expert assignments at each token during the forward propagation. Subsequently, we apply the nPMI measure defined in Eq.~\ref{eq:npmi} to identify the top five experts that exhibit the highest statistical dependence on reasoning-related indicators, such as the ``\texttt{<think>}'' token. These selected experts are thus regarded as the core cognitive components specialized in mathematical reasoning. Table~\ref{tab:expert_of_qwen} demonstrates the cognitive experts across math, physics, chemistry, and biology domain.

\begin{table*}[!t]
\caption{
Identified cognitive experts of Qwen3-235B. Each entry (layer ID, expert ID) denotes the Qwen3-235B model layer ID and expert ID. 
``All'' combines data from all domains.
}
\centering
\begin{tabular}{c ccccc} 
\toprule
\multirow{2}{*}{\bf Domain}  &   \multicolumn{5}{c}{\bf Identified Experts}\\
\cmidrule(lr){2-6}
    &   \em Top-1   &   \em Top-2    &   \em Top-3   &   \em Top-4    &   \em Top-5\\
\midrule
Math        &   (39, 182)   &   (29, 126)    & (14, 114)     & (27, 45)      &  (16, 129) \\
Physics     &   (2, 28)   &   (74, 65)    & (4, 44)      & (25, 103)      &  (7, 36) \\
Chemistry   &   (32, 58)    &   (26, 30)    & (68, 35)      & (37, 57)     &  (25, 103)\\
Biology     &   (2, 28)    &   (26, 30)    & (67, 15)     & (82, 29)      &  (25, 103)  \\
\hdashline
All         &   (25, 103)    & (26, 30)     & (82, 29)     & (67, 15)      &  (37, 57) \\
\bottomrule
\end{tabular}
\label{tab:expert_of_qwen}
\end{table*}

\begin{table*}[ht]
\centering
\caption{Pass@k performance of our cognitive experts on Deepseek-R1 and Qwen3-235B-A22B. For each problem, we generated 16 responses with a temperature of 0.6 and a top p value of 0.95.}
\begin{tabular}{l lccc}
\toprule
\textbf{Benchmark} & \textbf{Method} & \textbf{pass@1} & \textbf{pass@8} & \textbf{\#Tokens} \\
\midrule
\multirow{2}{*}{\textbf{AIME24}}
 & DeepSeek-R1  & 74.8 & 88.3 & 8,822 \\
 & ~~~+\method{} \{(39,182), (29,126)\} & 76.0 & 89.2  & 9,001 \\
\hdashline
\multirow{2}{*}{\textbf{AIME25}}
 & DeepSeek-R1  & 68.5 & 84.7  & 10,875 \\
 & ~~~+\method{} \{(39,182), (29,126)\} & 67.7 & 86.3  & 11,294 \\
\midrule
\multirow{2}{*}{\textbf{AIME24}}
 & Qwen3-235B       & 84.0 & 93.0  & 10,946\\
 & ~~~+\method{} \{(70,47), (23,115)\} & 85.0 & 91.6  & 10,706 \\
 \hdashline
\multirow{2}{*}{\textbf{AIME25}}
 & Qwen3-235B       & 82.7 & 88.3   & 12,546 \\
 & ~~~+\method{} \{(70,47), (23,115)\} & 82.1 & 89.7  & 12,373 \\
\bottomrule
\end{tabular}
\label{tab:passk}
\end{table*}

\subsection{Pass@k performance of cognitive experts}
\label{appendix:topp_math}
Table~\ref{tab:passk} presents the Pass@$k$ performance of our cognitive expert modulation approach compared to vanilla baselines across two model architectures. 
On DeepSeek-R1, our method demonstrates consistent improvements in Pass@8 accuracy (+0.9\% on AIME24 and +1.6\% on AIME25) despite showing marginal variations in Pass@1 performance. 
For Qwen3-235B-A22B, our approach achieves higher Pass@1 accuracy (+1.0\% on AIME24) while showing competitive Pass@8 performance (±1.4\% on AIME25), with consistent reductions in computational cost (2.2\% fewer tokens on AIME24 and 1.4\% fewer on AIME25). 

Under the Pass@1 metric \citep{guo2025deepseek}, the cognitive experts identified in AIME24 exhibit limited generalization to AIME25. This may be attributed to the top-p sampling mechanism, which partially dilutes the effectiveness of our approach, leading to average performance. In contrast, our method demonstrates strong generalization under greedy sampling, improving AIME25 accuracy from 63.3\% to 73.3\% on DeepSeek-R1 and from 66.7\% to 73.3\% on Qwen3-235B (in Table \ref{tab:expertVSself}). Thus, our approach enables correct answers in a single sampling step, eliminating the need for extensive sampling and verification, thereby enhancing sampling efficiency.

\subsection{Renormalization}
\label{appendix:Renormalization}

We investigate the DeepSeek Mixture-of-Experts (MoE) architecture, where each token selects 8 of 256 experts, with weights normalized to sum to 1. We examine steering specific expert weights under two conditions: with and without renormalization. The effects of the steering coefficient (reinforce factor) are presented in Table~\ref{tabl:normal}, with generalization performance analyzed in Table~\ref{tab:normal-generalization}.

Table~\ref{tabl:normal} evaluates the reinforce factor's effect on two cognitive experts. Without renormalization, accuracy peaks at 83.3\% (factors 4, 32, 64, 128) but drops to 3.3\% at 2048, with erratic token counts (e.g., 16,836). With renormalization, accuracy remains stable (73.3\%--83.3\%) across most factors, with token counts varying moderately (8,383--9,508), though it declines to 66.7\% at factor 256. Renormalization thus enhances robustness at higher steering coefficients.

We evaluate the generalization performance of cognitive experts, identified using normalized Pointwise Mutual Information (nPMI) within Mixture-of-Experts (MoE)-based large reasoning models, comparing three strategies: Vanilla R1, Renormalized, and Without Renormalized (wo/Renormalized). Table~\ref{tab:normal-generalization} reports performance across AIME25, Physics, Chemistry, Biology, and their average for experts selected from AIME24.

The wo/Renormalized strategy demonstrates superior generalization, achieving an average score of 73.1, compared to 70.9 for Vanilla R1 and 68.8 for Renormalized. This 4.3-point improvement over Renormalized is driven by notable gains in AIME25 (73.3 vs. 63.3) and Biology (79.0 vs. 68.4). In Physics, Vanilla R1 (91.9) outperforms wo/Renormalized (89.5, -2.4), while in Chemistry, Renormalized (52.7) surpasses wo/Renormalized (50.4), indicating domain-specific trade-offs.


\begin{table*}[t]
\caption{Reinforce factor effects of two cognitive experts with/without renormalization}
\centering
\begin{tabular}{rcccc}
\toprule
\multirow{2}{*}{\textbf{Reinforce Factor} } & 
\multicolumn{2}{c}{\textbf{wo/Renormalization}} & 
\multicolumn{2}{c}{\textbf{Renormalization}} \\
\cmidrule(lr){2-3} \cmidrule(lr){4-5}
& \em Acc & \em Token & \em Acc & \em Token\\
\midrule
1 (R1) & 73.3 & 9,291 & 73.3 & 9,291 \\
\hdashline
2         & 70.0 & 9,103 & 80.0 & 8,463 \\
4         & 83.3 & 8,145 & 80.0 & 8,383 \\
8         & 73.3 & 9,502 & 70.0 & 8,818 \\
16        & 80.0 & 8,493 & 73.3 & 9,133 \\
32        & 83.3 & 8,337 & 83.3 & 8,956 \\
64        & 83.3 & 8,317 & 80.0 & 9,508 \\
128       & 83.3 & 9,490 & 73.3 & 9,091 \\
256       & 60.0 & 7,986 & 66.7 & 8,719 \\
512       & 46.7 & 6,270 & 80.0 & 8,786 \\
1024      & 23.3 & 4,378 & 73.3 & 8,564 \\
\bottomrule
\end{tabular}
\label{tabl:normal}
\end{table*}

\begin{table*}[t]
\centering
\caption{Generalization capacity of two cognitive experts selected from AIME24, with or without renormalization.}
\begin{tabular}{lccccc}
\toprule
\textbf{Strategy} & \textbf{AIME25} & \textbf{Physics} & \textbf{Chemistry} & \textbf{Biology} & \textbf{Average} \\
\midrule
Vanilla R1       & 63.3 & 91.9 & 49.5 & 79.0 & 70.9 \\
Renormalized     & 63.3 & 90.7 & 52.7 & 68.4 & 68.8 \\
wo/Renormalized  & 73.3 & 89.5 & 50.4 & 79.0 & \bf 73.1 \\
\bottomrule
\end{tabular}
\label{tab:normal-generalization}
\end{table*}


\end{document}